\documentclass{egpubl}
\usepackage{pg2025}

\SpecialIssueSubmission    

\usepackage[T1]{fontenc}
\usepackage{dfadobe}  
\usepackage{cite}  
\BibtexOrBiblatex
\electronicVersion
\PrintedOrElectronic
\ifpdf \usepackage[pdftex]{graphicx} \pdfcompresslevel=9
\else \usepackage[dvips]{graphicx} \fi

\usepackage{egweblnk} 

\title[TIDE: Achieving Balanced Subject-Driven Image Generation via Target-Instructed Diffusion Enhancement]%
      {TIDE: Achieving Balanced Subject-Driven Image Generation via Target-Instructed Diffusion Enhancement}

\author[Jibai Lin \& Bo Ma et al.]
{\parbox{\textwidth}{\centering Jibai Lin$^{1,2,3}$\orcid{0009-0001-4722-4461}
         Bo Ma$^{1,2,3}$\orcid{0000-0003-3082-648X}
         Yating Yang$^{1,2,3}$\orcid{0000-0002-2639-3944}
         Xi Zhou$^{1,2,3}$
         Rong Ma$^{1,2,3}$
         Turghun Osman$^{1,2,3}$ 
         Ahtamjan Ahmat$^{1,2,3}$ \\
         Rui Dong$^{1,2,3}$ 
         Lei Wang$^{1,2,3}$
         }
         \vspace{-0.5em}
         \\
\parbox{\textwidth}{\centering $^1$Xinjiang Technical Institute of Physics \& Chemistry, Chinese Academy of Sciences\\
         $^2$University of Chinese Academy of Sciences\\
         $^3$Xinjiang Laboratory of Minority Speech and Language Information Processing
        }
}
\vspace{-8em}

\usepackage{amsmath}
\usepackage{amssymb}
\usepackage{mathtools}
\usepackage{booktabs} 
\usepackage{siunitx}  
\usepackage{caption}  
\usepackage{makecell} 
\usepackage{graphicx} 
\usepackage{multirow} 
\usepackage{array}    
\usepackage{xcolor}   
\usepackage{float}

\begin{document}

 \teaser{
  \includegraphics[width=0.75\linewidth]{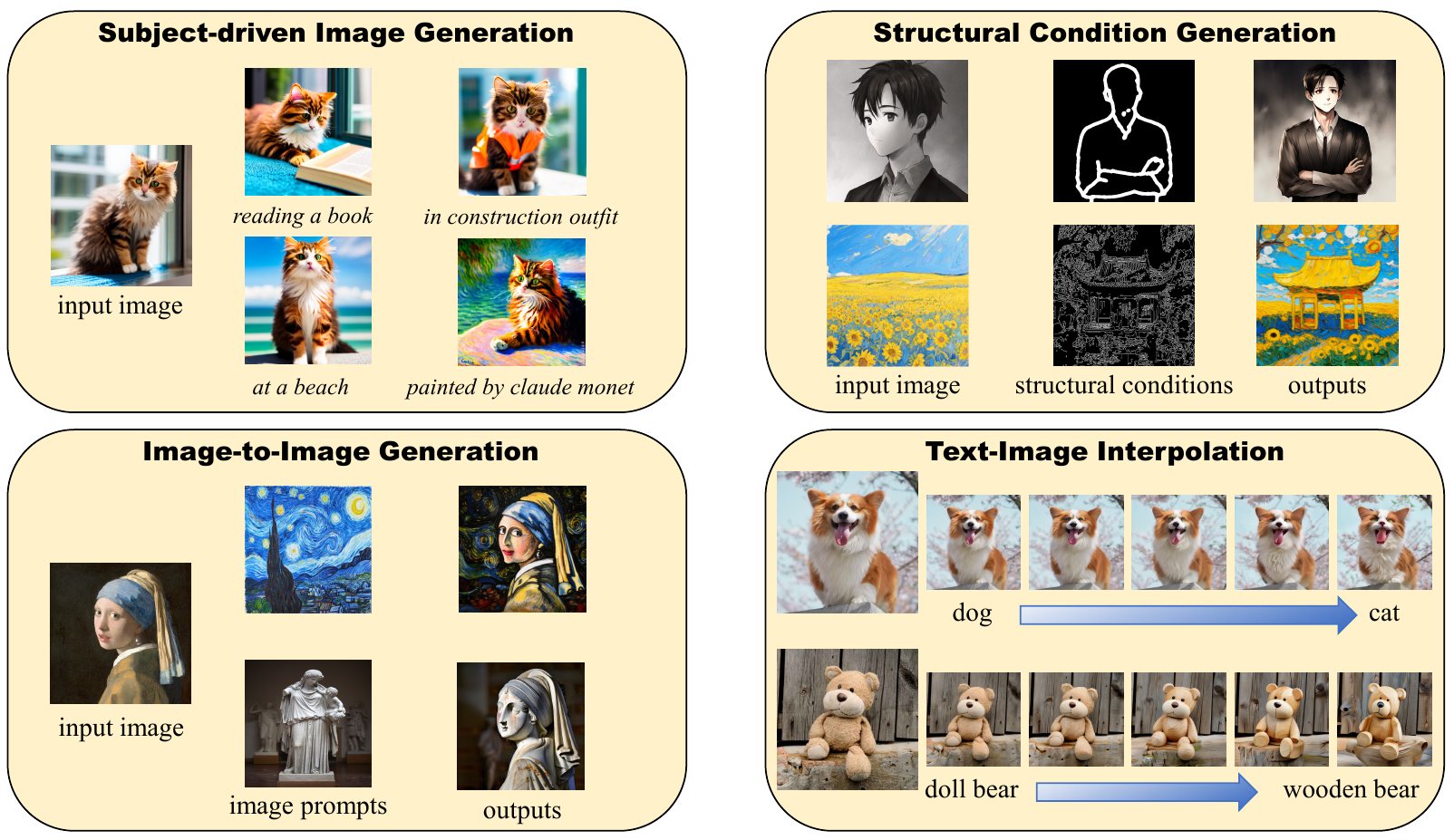}
  \centering
   \caption{\label{fig:ex1}Our TIDE framework achieves high-fidelity subject-driven generation from single reference images 
                           via target-supervised learning. Our model also supports diverse tasks like structure-controlled 
                           generation, image-to-image generation, and interpolation between reference images
                            and textual prompts to demonstrate its versatility in subject manipulation.}
 \label{fig:teaser}
}

\maketitle
\begin{abstract}

  Subject-driven image generation (SDIG) aims to manipulate specific subjects within images while 
  adhering to textual instructions, a task crucial for advancing text-to-image diffusion models. 
  SDIG requires reconciling the tension between maintaining subject identity and complying with dynamic 
  edit instructions, a challenge inadequately addressed by existing methods. In this paper, we introduce 
  the Target-Instructed Diffusion Enhancing (TIDE) framework, which resolves this tension through target 
  supervision and preference learning without test-time fine-tuning. TIDE pioneers target-supervised 
  triplet alignment, modelling subject adaptation dynamics using a (reference image, instruction, target 
  images) triplet. This approach leverages the Direct Subject Diffusion (DSD) objective, training the 
  model with paired "winning" (balanced preservation-compliance) and "losing" (distorted) targets, 
  systematically generated and evaluated via quantitative metrics. This enables implicit reward modelling 
  for optimal preservation-compliance balance. Experimental results on standard benchmarks demonstrate 
  TIDE's superior performance in generating subject-faithful outputs while maintaining instruction 
  compliance, outperforming baseline methods across multiple quantitative metrics. TIDE's versatility 
  is further evidenced by its successful application to diverse tasks, including structural-conditioned 
  generation, image-to-image generation, and text-image interpolation. Our code is available at \url{https://github.com/KomJay520/TIDE}.

\end{abstract}  
\section{Introduction}

Subject-Driven image generation (SDIG) aims to synthesize images that preserve the identity
of a specified subject (e.g., a cat, object) while adapting to textual editing instructions 
(e.g., "The cat* swimming in a pool")\ \cite{gal2022image, ruiz2023dreambooth, kumari2023multi, li2023blip, zhang2024ssr}. 
This task extends text-to-image diffusion models\ \cite{dhariwal2021diffusion, rombach2022high, ahn2024self, chefer2024hidden} to enable controlled subject manipulation, 
critically expanding their utility for realistic applications like personalized content creation 
and visual storytelling. At its core, SDIG demands precise reconciliation of two competing 
objectives: invariant preservation of subject identity against dynamic compliance with edit 
instructions. This process advances controllable generation techniques for real-world subject editing scenarios.

\begin{figure}[h]
  \centering
  \includegraphics[width=\linewidth]{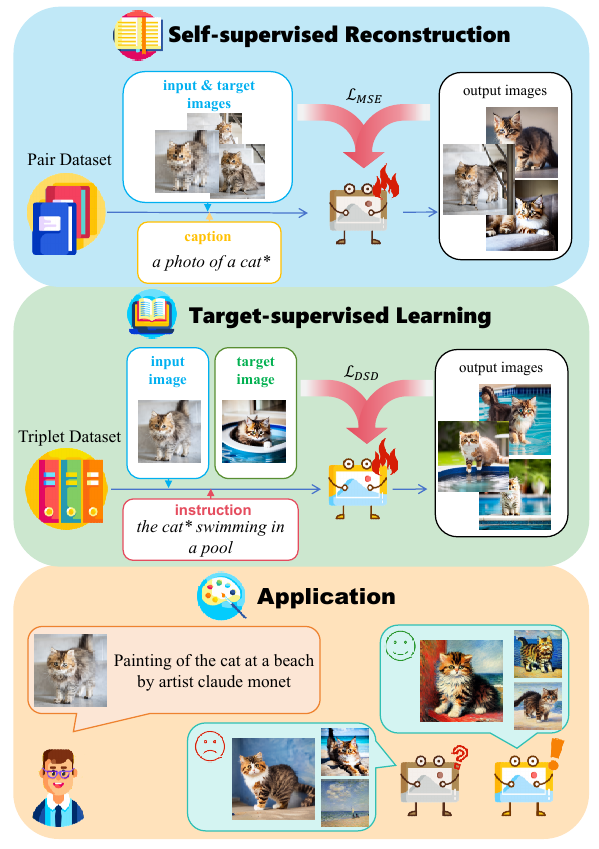}
  \caption{\label{fig:ex2}Different of Self-supervised Reconstruction and Target-supervised Learning in subject-driven image generalization. 
  Self-supervised Reconstruction always use the input images as label images, result to bad output in applications.}
\end{figure}

Existing SDIG methods can be broadly classified into fine-tuning based and fine-tuning free methods, 
with the key distinction lying in whether they perform parameter updates during test time. 
On the one hand, fine-tuning based methods fine-tune pre-trained diffusion model or its components
using 3-5 reference images at test-time\ \cite{ruiz2023dreambooth, avrahami2023break, hyung2024magicapture}. 
These methods embed subject features into a placeholder token (e.g., [V]) and combine it with 
text prompts (e.g., The [V] cat swimming in a pool) to generate subject-driven images. However, they suffer from catastrophic forgetting: 
retraining on novel subjects leads to the overwriting of previously learned representations 
limiting their scalability to diverse subject domains. On the other hand, fine-tuning free methods 
preserve parameters of base model by encoding reference images via external modules (CLIP/BLIP) and injecting 
subject features through trainable adapters\ \cite{li2023blip, zhang2024ssr, ma2024subject}. Yet, their self-supervised reconstruction paradigm, as shown 
in the top row of Figure~\ref{fig:ex2}, is forced to use reference images as pseudo-targets. Due to missing 
instruction-edited pairs, they conflate preservation (e.g., retaining object appearance) and modification 
(e.g.,altering background). This ambiguity prevents models from learning the critical balance between 
feature retention and semantic adaptation. In summary, existing approaches fail to inadequate 
balance subject preservation and instruction compliance: fine-tuning based methods sacrifice generalization 
for subject fidelity, while fine-tuning free methods lack explicit supervision for edit disentanglement.

To resolve the preservation-compliance trade-off, we propose TIDE (Target-Instructed Diffusion Enhancing), 
a fine-tuning-free framework built upon two key innovations: explicit triplet supervision and 
the Direct Subject Diffusion (DSD) objective. First, as demonstrated in the middle row of Figure~\ref{fig:ex2}, 
TIDE introduces reference-instruction-target triplet supervision, which is a structured guidance mechanism 
absent in prior self-supervised methods. Unlike approaches that rely solely on reference images as targets (top row), 
the target images in triplet explicitly encode desired attribute edits (e.g., pose, background changes), 
enabling the model to learn fine-grained mappings between instructions and subject adaptations. Crucially, 
TIDE proposes the DSD objective tailored for SDIG. Specifically, DSD adapts the pairwise preference learning 
to image generation by integrating subject-specific features from reference images, enabling the model to 
learn from paired "winning"/"losing" target images. For each instruction (e.g., "Cat swimming in a pool"), the preference pairs which encode human-like 
quality judgments for balanced outputs, train the model to prefer "winning" targets (balanced outputs) over 
"losing" targets (unbalanced, e.g., instruction or feature loss). This preference signal allows TIDE to 
implicitly model a reward function that guides the diffusion process to generate semantically consistent 
images while preserving critical subject attributes. As visualized in Figure~\ref{fig:ex2} bottom-row 
comparisons, TIDE generates images faithfully following instructions while preserving subject identity.

To implement these innovations, we adapt prior multimodal fusion techniques through an Image Projection Module 
with Image Cross-Attention Module (IPM-ICAM). This module encodes reference features to condition the diffusion process. 
Extensive experiments have shown that our method with this lightweight module achieves 
comparable results to other methods in subject-driven image generation.

To summarize, our contributions are listed as follows:
\begin{itemize}
  \item We propose a novel framework, named TIDE, for personalized subject-driven image generation, 
    which uses triplet data to explicitly specify feature preservation and modification rules.
  \item In TIDE, we propose Direct Subject Diffusion, a novel preference learning objective that enables 
    implicit optimization of both subject fidelity and instruction compliance. 
  \item Our extensive experiments have validated the robustness and flexibility of our approach, showcasing its 
    capability to deliver competitive performance among SDIG methods. It 
    also supports diverse tasks including structural-conditioned generation, image-to-image generation, and text-image interpolation.
\end{itemize}

\section{Related Work}

\subsection{Aligning Large Language Models}

LLMs\ \cite{achiam2023gpt, ye2024mplug} are typically aligned with human preferences through supervised fine-tuning (SFT)\ \cite{dong2023abilities} on 
demonstration data followed by reinforcement learning from human feedback (RLHF)\ \cite{bai2022constitutional}. RLHF learns a reward model from pairwise output comparisons to guide policy 
optimization via policy-gradient methods\ \cite{mnih2016asynchronous}, but faces notable challenges including high 
computational costs from nested training loops, sensitivity to hyper-parameter 
choices, and susceptibility to reward hacking\ \cite{dubois2023alpacafarm, skalse2022defining}. Recent approaches mitigate 
these issues through reward-weighted supervised learning, ranking loss 
minimization and direct policy optimization\ \cite{krishnaiah2024harmonizing, rafailov2023direct, dubois2023alpacafarm}. The latter matches the 
performance of RLHF without its complexity. Direct Preference Optimization (DPO) has since been widely adopted to bridge structural 
discrepancies in various domains. Inspired by DPO\ \cite{rafailov2023direct, wallace2024diffusion}, a framework that transforms RLHF into a 
supervised learning problem, our method leverages its core principles to address subject-driven alignment challenges. 
Promising extensions using AI feedback\ \cite{bai2022constitutional} suggest pathways 
for scalable alignment in SDIG.

\subsection{Text-to-image models}

Recent advancements in large-scale diffusion models have significantly expanded their 
applicability in image synthesis\ \cite{ramesh2021zero, ramesh2022hierarchical, rombach2022high, podell2023sdxl, lin2024accdiffusion}. Among these, latent diffusion models (LDMs)\ \cite{rombach2022high} have emerged 
particularly noteworthy due to their computational efficiency. The evolution of text-to-image 
generation began with DALLE\ \cite{ramesh2021zero} which introduces the autoregressive transformer architecture, followed by 
innovative integration of a diffusion prior module with cascaded super-resolution decoders in DALLE-2\ \cite{ramesh2022hierarchical}. 
Concurrently, Imagen\ \cite{saharia2022photorealistic} demonstrated the effectiveness 
of large pretrained language models (T5)\ \cite{raffel2020exploring} for enhanced textual understanding combined with 
large-scale diffusion training. DeepFloyd IF\ \cite{saharia2022photorealistic} further advanced the field through a three-stage 
cascaded diffusion architecture capable of generating legible typography. Stable Diffusion (SD)\ \cite{rombach2022high} established itself as a benchmark 
by introducing cross-attention mechanisms for precise text-conditioned generation, combining high-quality output with 
computational practicality. Our framework builds upon Stable Diffusion, selected for its modular design and open-source 
availability, features that facilitate extensible adaptation.

\subsection{Subject-driven Diffusion models}

The widespread adoption of diffusion models for text-to-image synthesis has naturally 
extended to subject-driven generation tasks\ \cite{gal2022image, ruiz2023dreambooth, wei2023elite, ye2023ip, li2023blip, zhang2024ssr, ma2024subject}. There are two main frameworks for customized 
subject-driven image generation from the perspective of whether to introduce test-time fine-tuning.

Fine-tuning based methods often optimize additional text embeddings or directly 
fine-tune the diffusion model to fit the desired subject \ \cite{gal2022image, ruiz2023dreambooth, kumari2023multi}. For instance, Textual Inversion\ \cite{gal2022image} 
aligns subject image with additional text embeddings, while DreamBooth\ \cite{ruiz2023dreambooth} adjusts the entire 
U-net in the diffusion model. Other methods like Custom Diffusion\ \cite{kumari2023multi} and Cones\ \cite{liu2023cones} only fine-tunes 
the K and V layers of the cross-attention. SVDiff\ \cite{han2023svdiff} fine-tunes the singular values of the 
weight matrices. These methods both try to minimize the parameters needing finetuning and 
reduce computational demands. On the other hand, after Custom Diffusion\ \cite{kumari2023multi} proposes the 
personalized generation of multiple subjects for the first time, Cones2\ \cite{liu2023cones2} generates two-subject 
combination images by learning the residual of token embedding and controlling the attention 
map. MagiCapture\ \cite{hyung2024magicapture} introduces a multi-concept personalizing method capable of generating 
high-resolution portrait images that faithfully capture the characteristics of both source 
and reference images. Perfusion\ \cite{tewel2023key} develop a gated rank-1 approach that enables us to control 
the influence of a learned concept during inference time and to combine multiple concepts.

Although fine-tuning based methods have better subject guidance ability, they suffer from 
a catastrophic forgetting problem and ineffective usage problem (e.g., given a new 
subject, these methods require complete model retraining, leading to efficiency bottlenecks), 
another research route involves constructing a large amount of domain-specific data or using 
open-domain image data for training without additional fine-tuning. Fine-tuning-free 
methods typically train an additional structure to encode the reference image into 
embeddings or image prompts\ \cite{gal2023encoder, ye2023ip, ma2024subject}. E4T\ \cite{gal2023encoder} uses a set of weight-offsets for the text-to-image model 
that learn how to effectively ingest additional concepts. ELITE\ \cite{wei2023elite} employs global and local 
mapping trained on OpenImages\ \cite{kuznetsova2020open}, but achieves limited text alignment due to architectural 
constraints. InstantBooth\ \cite{shi2024instantbooth} proposes an adapter structure inserted in the U-net. 
UMM-Diffusion\ \cite{ma2023unified} introduces a noval Unified Multi-Modal Latent Diffusion that processes joint 
text-image inputs containing target subjects to generate customized outputs. However, its 
reliance on the LAION-400M\ \cite{schuhmann2021laion} dataset limits performance for rare themes. Similarly, 
domain-specific training data is employed by E4T, InstantBooth, FastComposer\ \cite{xiao2024fastcomposer}, and Face-Diffuser\ \cite{wang2024high}. 
BLIP-Diffusion\ \cite{li2023blip} propose two-stage training scheme to achieve good fidelity effects. IP-Adapter\ \cite{ye2023ip} encodes images into prompts whit a 
lightweight adapter and decoupled cross-attention strategy. Subject-Diffusion\ \cite{ma2024subject}, SSR-Encoder\ \cite{zhang2024ssr} 
and FreeCustom\ \cite{ding2024freecustom} use segmentation network to mask the key concept from visual input, improving 
subject consistence in multi-subjects-driven generation. These methods still rely on 
self-supervised reconstruction objectives, resulting in failure to maintain subject consistency 
under complex editing instructions. Specifically, this limitation stems from their training paradigm, 
which uses input images as reconstruction targets and captions as proxy instructions, 
ultimately limiting their adaptability to diverse editing commands. 
These critical limitations motivate our work's key innovations.


\begin{figure*}[tbp]
  \centering
  \mbox{} \hfill
  \includegraphics[width=\linewidth]{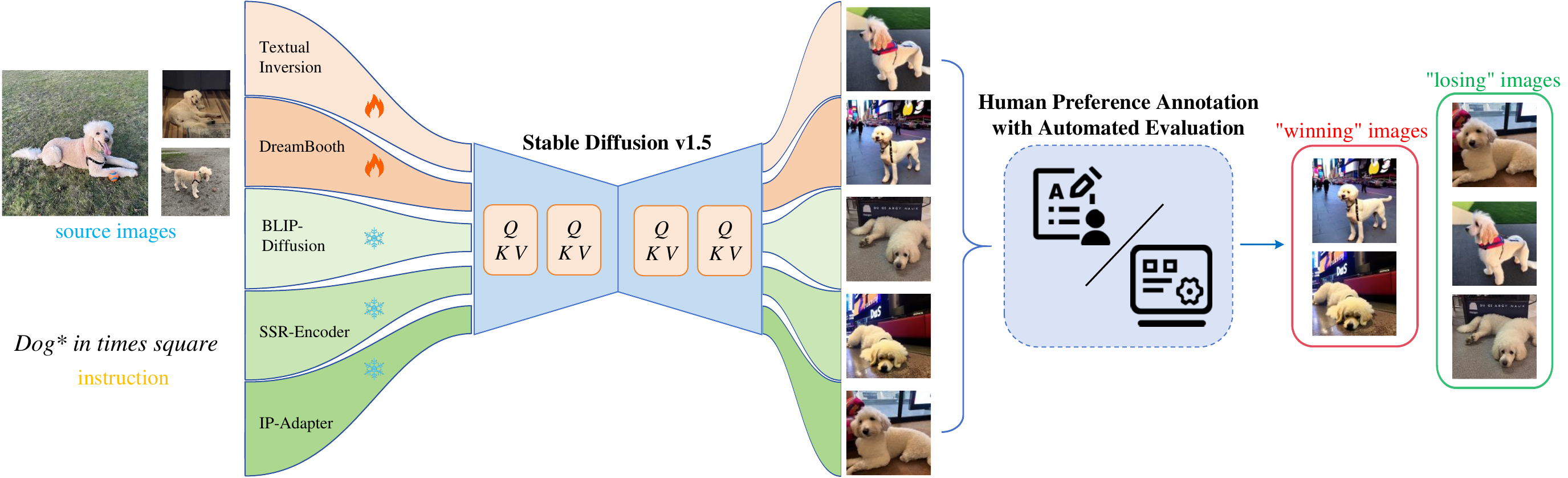}
  \hfill \mbox{}
  \caption{\label{fig:ex4}%
           The procedure for training data generation. (i) Samples are first generated using different baseline methods 
           with the same diffusion model. Among them, Textual Inversion and DreamBooth require fine-tuning the model, 
           while others do not. (ii) Using automated evaluation models (CLIP, DINO), generated samples are sorted: level 
           5-4 samples are selected as "winning" samples, and the rest as "losing" samples.}
\end{figure*}

\section{Preliminaries}

\subsection{Diffusion Models}

Diffusion models (DMs) learn to generate samples by progressively denoising a Gaussian-distributed 
variable, starting from data samples drawn from the distribution $q(x_{0})$. They are trained to 
reverse diffusion processes that reconstruct clean data from noisy inputs. The forward diffusion 
process is defined as:
\begin{equation}
    x_{t} =\sqrt{1-\beta _{t} }x_{t-1}+\sqrt{\beta _{t} }n_{t}, \qquad t=1,...,T \label{eq:eq1}
\end{equation}
where $x_0$ is the source image from dataset, $\beta _{t}$ denotes the noise schedule and $n_{t}$ represents independent and identically
distributed (i.i.d.) Gaussian noise vectors. Through mathematical derivation, the process can be
equivalently expressed as: 
\begin{equation}
    x_{t} =\sqrt{\bar{\alpha} _{t}}x_{0}+\sqrt{1-\bar{\alpha}_{t} } \epsilon _{t} \label{eq:eq2}
\end{equation}
where $\alpha _{t}=1-\beta _{t}, \bar{\alpha} _{t} =  {\textstyle \prod_{i=1}^{t}}\alpha _{i}$ and 
$\epsilon _{t}\sim \mathcal{N}(0, \mathrm {I})$ is random Gaussian noise. We use this algorithm to inverse our target images. Then to generate 
a new image $\hat{x}_{0} $ following the deterministic DDIM\ \cite{song2020denoising}, the reverse diffusion process starts from 
a random noise $\hat{x}_{T}\sim \mathcal{N}(0, \mathrm {I})$, which can be iteratively denoised as: 
\begin{align}
    \hat{x}_{t-1}= & \sqrt{\bar{\alpha}_{t-1}}\frac{\hat{x}_{t}-\sqrt{1-\bar{\alpha}_{t}}\epsilon_{\theta}(\hat{x}_{t})}{\sqrt{\bar{\alpha}_{t}}} \nonumber\\
    &+\sqrt{1-\bar{\alpha}_{t-1}}\epsilon_{\theta}(\hat{x}_{t})   , \qquad t=T,...,1 \label{eq:eq3}
\end{align}
Here $\epsilon_{\theta}(\cdot)$ is an estimate of $\epsilon _{t}$ produced by our frozen neural 
network DM and a trainable adapter model with learned parameters $\theta$. For SDIG, the 
model is conditioned on a text prompt $p$ and subject images $I$ to produce images faithful to these. 
Instead of estimating the loss between image outputs and target outputs, DMs' training is performed 
by minimizing the evidence lower bound (ELBO)\ \cite{kingma2021variational}:
\begin{equation}
    L_{DM} = \mathbb{E}_{\hat{x}_{0},\epsilon,t,\hat{x}_{t}} \left[ \omega(\lambda_{t}) \, \left \| \epsilon - \epsilon_{\theta}(\hat{x}_{t}, c_{p}, t) \right\| _{2}^{2} \right] \label{eq:eq4}
\end{equation}
where $\lambda_{t}$ is a signal-to-noise ratio, $\omega(\lambda_{t})$ is a pre-specified weighting 
function which typically keeps constant, timestep $t\sim \mathcal{U}(0,T)$, $c_{p}$ is the condition (or feature) from $p$.

\subsection{DPO for Diffusion Models}

RLHF aims to optimize a conditional distribution $p_{\theta}(y_{0}|c)$ 
(conditioned on $c\sim \mathcal{D}_c$, $y_{0}$ is target output) such that the latent reward model
$r(c,y_{0})$ define on it is maximized, while regularizing the KL-divergence from a reference 
distribution $p_{ref}$:
\begin{align}
\max_{p_{\theta}}\mathbb{E} & _{c\sim\mathcal{D}_c,y_{0}\sim p_{\theta}(y_{0}|c)}[r(c,y_{0})] \nonumber \\
& -\beta \mathbb{D}_{\mathrm {KL}}[p_{\theta }(y_{0}|c)||p_{ref}(y_{0}|c)]  \label{eq:eq5}
\end{align}
where the hyperparameter $\beta$ controls regularization.

In DPO\ \cite{rafailov2023direct}, we always assume access only to ranked pairs generated from some conditioning 
$y_{0}^{w}\succ y_{0}^{l} |c$, where $y_{0}^{w}$ and $y_{0}^{l}$ denote the "winning" (positive) 
and "losing" (negative) samples. With the loss function of Bradley-Terry (BT) model\ \cite{imrey2005b}, the loss 
function is estimated via maximum likelihood training for binary classification:
\begin{align}
L_{\mathrm {BT} }(\phi )=-\mathbb{E}_{c, y_{0}^{w}, y_{0}^{l}}[log \sigma (r_{\phi}(c,y_{0}^{w})-r_{\phi}(c,y_{0}^{l}))] \label{eq:eq6}
\end{align}
where $\sigma$ is the sigmoid function. $r_{\phi}(\cdot)$ is reward function which is parameterized 
by a neural network $\phi$. Using Eq.\eqref{eq:eq5}, this leads to the DPO objective:
\begin{align}
L_{\mathrm {DPO}}(\theta)=-\mathbb{E}_{c,y_{0}^{w},y_{0}^{l}}\left [  log\sigma \left (\beta log\frac{p_{\theta }(y_{0}^{w}|c)}{p_{\mathrm {ref}(y_{0}^{w}|c)}}-\beta log\frac{p_{\theta }(y_{0}^{l}|c)}{p_{\mathrm {ref}(y_{0}^{l}|c)}}\right )\right ] \label{eq:eq7}
\end{align}
By combining Eq.\eqref{eq:eq4} and Eq.\eqref{eq:eq7}, DPO-Diffusion (DD)\ \cite{wallace2024diffusion} adapt DPO for diffusion model yields:
\begin{align}
L_{\mathrm{DD}}(\theta ) & = -\mathbb{E}_{(p,y_{0}^{w},y_{0}^{l})\sim \mathcal{D},t\sim\mathcal{U}(0,T), y_{t}^{w}\sim q(y_{t}^{w}|y_{0}^{w}),y_{t}^{l}\sim q(y_{t}^{l}|y_{0}^{l})} \nonumber \\
   & \qquad \qquad \quad  log\sigma (-\beta T\omega (\lambda _{t})( \nonumber \\
   & \left \| \epsilon ^{w}-\epsilon _{\theta }(y_{t}^{w},c_{p},t) \right \| _{2}^{2}-\left \| \epsilon ^{w}-\epsilon _{\mathrm {ref}   }(y_{t}^{w},c_{p},t) \right \| _{2}^{2} \nonumber\\
   & -(\left \| \epsilon ^{l}-\epsilon _{\theta }(y_{t}^{l},c_{p},t) \right \| _{2}^{2}-\left \| \epsilon ^{l}-\epsilon _{\mathrm {ref}   }(y_{t}^{l},c_{p},t) \right \| _{2}^{2}))) \label{eq:eq8}
\end{align}
where $q(y_{t}^{w}|y_{0}^{w})$ is the diffusion process following Eq.\eqref{eq:eq2}, 
$\epsilon _{\mathrm {ref}}(\cdot)$ denotes predictions from the frozen reference model. 

\section{Methodology}

In this section, we first present our open-domain instruction-target dataset for supervised 
training in SDIG. Next, we show an overview of the TIDE framework, 
followed by an explanation of how we leverage multimodal information to train external adapter, 
including multimodal representation and Direct Subject Diffusion optimization.

\subsection{Dataset Construction} \label{sec:data}

To enable diffusion models to balance subject preservation and instruction compliance, a 
multimodal supervised dataset with open-domain coverage is essential. Existing 
datasets\ \cite{kuznetsova2020open, schuhmann2022laion}, primarily designed for multimodal alignment and visual question answering 
(e.g., image-caption pairs), prove inadequate for this purpose due to their inherent 
structural limitations. This motivates our creation of a tailored, high-quality 
dataset comprising three core components: (1) visual subject inputs, (2) structured 
textual instructions with fine-grained control parameters, and (3) corresponding 
target outputs optimized for subject-driven generation scenarios.

Figure~\ref{fig:ex4} illustrates our methodology for constructing the training dataset derived 
from Concept101\ \cite{kumari2023multi} benchmark, which provides diverse and semantically rich instructions per 
subject. We employ existing subject-driven image generation techniques to produce candidate 
outputs, and then use quantifiable quality metrics, the similarity of CLIP and DINOv2 embeddings, 
to select "winning" and "losing" sample pairs. Specifically, we calculate 
the similarity between the embeddings of the output samples and prompts. We use a 
hyperparameter $\varphi$ to balance the textual similarity and the visual similarity. 
For each candidate output, we compute two normalized similarity metrics: (1) 
$S_{text} = \mathrm{mean}\left(\mathrm {CLIP}(x_{p}) \cdot \mathrm {CLIP}(y_{0})^{\top}\right)$ for instruction alignment, 
and (2) $S_{visual} = cos(\mathrm {DINO}(x_{i}), \mathrm {DINO}(y_{0}))$ for subject 
alignment. The composite quality score $Q$ is then computed as:
\begin{align}
    Q = \varphi \cdot S_{text} + (1-\varphi)\cdot S_{visual} \label{eq:eq13}
\end{align}
where $\varphi \in [0,1]$ is a tunable hyperparameter that balances the relative 
importance of textual versus visual fidelity. Through extensive grid search, we set 
$\varphi=0.7$ to prioritize instruction following while maintaining adequate subject 
preservation.

\begin{figure}[h]
  \centering
  \includegraphics[width=\linewidth]{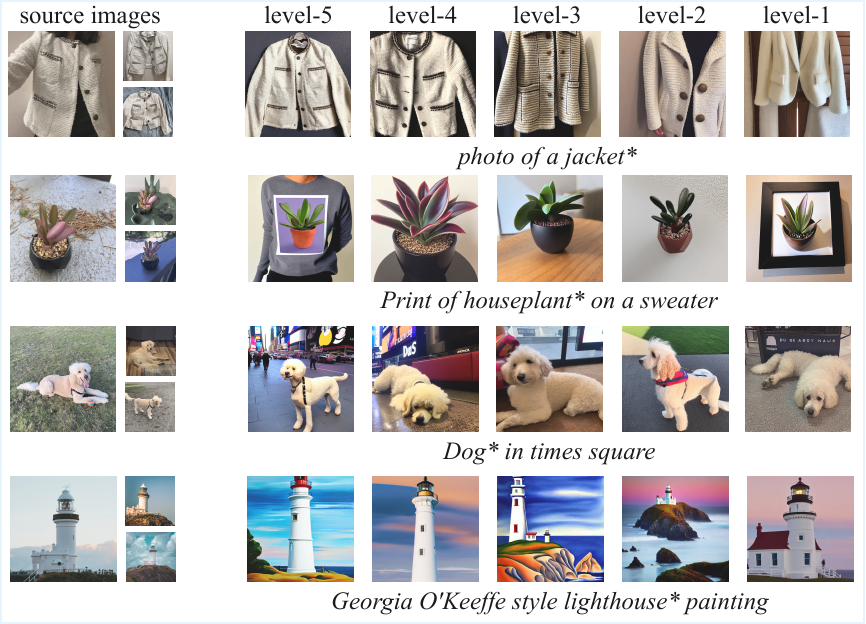}
  \caption{\label{fig:ex5}%
           Sample images from the dataset showing source and target 
           images with instructions. Level 5 means the best target image. 
           We choose the level 5--4 images as the winning samples.}
\end{figure}

Through this automated selection strategy, we establish C4DD (Concept101 for DPO Dataset). 
Image candidates are ranked into five quality levels (Level 5: highest $Q$), where Level 5--4 
images are designated as "winning" samples and Level 3--1 as "losing" samples (Figure~\ref{fig:ex5}). 
This dataset contains 10,000 high-quality samples with 6,000 carefully curated 
(winning, losing) pairs. To the best of our knowledge, C4DD is the first target-supervised dataset for 
SDIG.


\begin{figure*}[tbp]
  \centering
  \includegraphics[width=\linewidth]{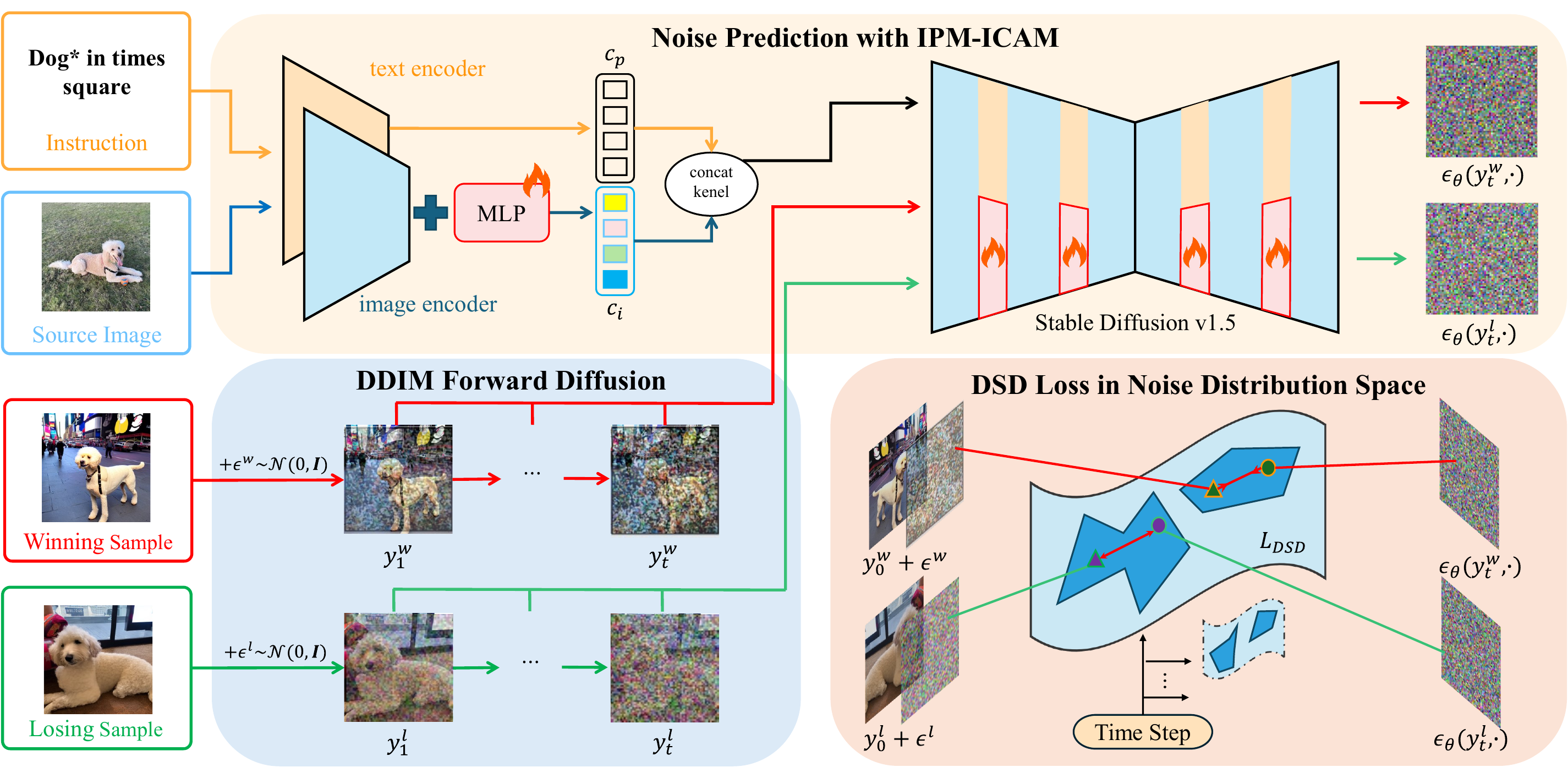}
  \hfill \mbox{}
  \caption{\label{fig:ex3}%
           Illustration of the DSD training for TIDE. (i) Given a query instruction-image pair, we first design IPM-ICAM 
           to align the image feature $c_i$ with instruction embedding $c_p$. (ii) Then, based on DDIM forward diffusion, we convert 
           target samples into noisy latent $(y_t^w, y_t^l)$ and their noise residuals $(\epsilon^w, \epsilon^l)$ at timestep $t$.
           (iii) After that, Stable Diffusion v1.5 predicts the target noise from noisy latent using fused multimodal feature $F(c_p, c_i)$. 
           (iv) Finally, the DSD loss compares predicted noise $(\epsilon_{\theta}(y_t^w, \cdot), \epsilon_{\theta}(y_t^l, \cdot))$ 
           with noise residuals $(\epsilon^w, \epsilon^l)$ in noise distribution space, learning preservation-compliance balance 
           $(\cdot$ denotes $F(c_p, c_i)$ and $t$).}
\end{figure*}

\subsection{Model Overview} \label{sec:model}
As illustrated in Figure~\ref{fig:ex3}, TIDE framework resolves the preservation-compliance 
trade-off through three coordinated design principles. First, the input pipeline 
processes triplets of instruction $x_p$, source image $x_i$ and "winning"/"losing" target images 
$(y_{0}^{w}, y_{0}^{l})$ from our C4DD. The target images establish explicit supervision for 
balanced generation. Second, the trainable adapter IPM-ICAM dynamically align source image feature $c_i$
with instruction feature $c_p$. And then 
the diffusion model predict the noise relying on this multimodal feature $F(c_p, c_i)$. Third, by leveraging the DDIM 
forward diffusion following the Eq.\eqref{eq:eq2}, we get the noise signals $\epsilon^{w}$ and $\epsilon^{l}$ 
from the target images. The DSD objective can operate in noise space by comparing predictions 
of $\epsilon^{w}$ and $\epsilon^{l}$. The loss optimizes the trainable adapter to concentrate on 
gaps between "winning" and "losing" targets through preference-aware gradients.

\subsection{Mulitimodal Representation}

Our framework builds on the native architecture of Stable Diffusion, which employs a frozen CLIP text 
encoder for prompt conditioning. To ensure modality consistency, we adopt the CLIP image encoder 
for subject representation extraction, leveraging their shared latent space to achieve coarse-grained 
preliminary visual-textual alignment essential for subsequent cross-modal fusion. This design 
preserves the original text-image correlation capabilities from base model while enabling synergistic 
cross-modal interaction.

As illustrated in Figure~\ref{fig:ex3}, we introduce a lightweight adaptation module 
comprising two key components: (1) Image Projection Module (IPM): A multilayer perceptron that maps 
CLIP visual features to the cross-attention dimension, eliminating the need for visual encoder 
fine-tuning. (2) Image Cross-Attention Module (ICAM): A direct adaptation of original text 
cross-attention mechanism used in SD. To formalize the interaction between these components, 
the following equations define their operations: 
\begin{align}
c_{i} & = \mathrm{IPM}(x_{i})=MLP(\mathrm{CLIP} (x_{i})) \label{eq:eq9} \\
z & = \mathrm {ICAM}(Q, K^{i}, V^{i}) = \mathrm {Softmax}(\frac{Q(K^{i})^{\top}}{\sqrt{d}})V^{i} \label{eq:eq10}
\end{align}
where $K^{i} = W^{k}\cdot c_{i}$ and $V^{i} = W^{v}\cdot c_{i}$, $W^{k}$ and $W^{v}$ are trainable 
parameters. 

Crucially, while the base diffusion model remains entirely frozen, only lightweight components 
of IPM-ICAM (merely 1.33\% of base model parameters) are updated during training. This preserves the 
model original generative capabilities while enabling controlled subject manipulation through the 
learned adapter.

\subsection{Direct Subject Diffusion Optimization}

Building upon the aligned multimodal representations, we propose Direct Subject Diffusion (DSD) objective, 
the first Direct Preference Optimization adaptation for SDIG. 
Unlike conventional DPO that operates solely on text prompts, DSD extends to visual subjects and textual instructions, encoding 
user preference via supervised winning/losing image pairs. This multimodal preference modeling framework enables fine-grained 
supervision based on the perceived quality of generated outputs without relying on per-sample reward engineering 
or direct loss formulations for subjective concepts.
Each training instance $ \left (x_{p}, x_{i}, y_{0}^{w}, y_{0}^{l} \in \mathcal{D} \right )$ conditions generation on both 
textual and visual inputs, where $y_{0}^{w}\succ y_{0}^{l}$ indicates human preference for preservation-compliance balance. 
To inject multimodal conditioning into the generation process, we leverage the IPM-ICAM adapter, which transforms and 
aligns visual and textual features into a shared representation. Specifically, the conditioning signal can be formalized as: 

\begin{equation}
    F(c_{p},c_{i})=\underbrace{\mathrm {Attention}(Q,K^{p},V^{p})}_{\text{Text Guidance}}+\gamma\cdot \underbrace{\mathrm{ICAM}(Q,K^{i},V^{i})}_{\text{Subject Anchoring}} \label{eq:eq11}
\end{equation}
The text guidance term interprets the instruction semantics and encapsulates high-level editing intent. 
The subject anchoring term reinforces the presence and identity of the visual subject extracted from the image. 
These two components are weighted by a hyperparameter $\gamma$  to allow flexible control over the influence of the visual cue, with 
$\gamma=0$ reducing the model to standard text-to-image diffusion. 

For each target pair, we apply DDIM forward diffusion to produce noisy latent variables $(y_{t}^{w}, y_{t}^{l})$, 
along with their corresponding noise residuals $\epsilon ^{w}$ and $\epsilon ^{l}$ at timestep $t$. The model, 
parameterized by $\theta$, is trained to predict the original noise from each noisy latent (i.e., $\epsilon _{\theta }(y_{t}^{w},F(c_{p},c_{i}),t)$). 
Finally, based on Eq.\eqref{eq:eq8} and Eq.\eqref{eq:eq11}, our proposed Direct Subject Diffusion (DSD) 
objective extends conventional DPO to handle the unique requirements of subject-conditioned generation:

\begin{align}
L_{\mathrm{DSD}}(\theta)& = -\mathbb{E}_{(p,i,y_{0}^{w},y_{0}^{l})\sim \mathcal{D},\; t\sim\mathcal{U}(0,T),\; y_{t}^{w}\sim q(y_{t}^{w}|y_{0}^{w}),\; y_{t}^{l}\sim q(y_{t}^{l}|y_{0}^{l})} \nonumber\\
& \log\sigma \bigg(-\beta T\omega (\lambda _{t}) \Big( \underbrace{\| \epsilon ^{w}-\epsilon _{\theta }(y_{t}^{w},F(c_{p},c_{i}),t) \|_{2}^{2}}_{\text{Preference Alignment}} \nonumber\\
& \qquad - \underbrace{\| \epsilon ^{w}-\epsilon _{\mathrm{ref}}(y_{t}^{w},F(c_{p},c_{i}),t) \|_{2}^{2}}_{\text{Reference Regularization}} \nonumber\\
& \qquad - \underbrace{\| \epsilon ^{l}-\epsilon _{\theta }(y_{t}^{l},F(c_{p},c_{i}),t) \|_{2}^{2}}_{\text{Dispreference Repulsion}} \nonumber\\
& \qquad + \underbrace{\| \epsilon ^{l}-\epsilon _{\mathrm{ref}}(y_{t}^{l},F(c_{p},c_{i}),t) \|_{2}^{2}}_{\text{Reference Normalization}}  \Big) \bigg) \label{eq:eq12}
\end{align}
Here $\epsilon _{\theta }$ denotes the noise prediction from trainable model, while $\epsilon _{\mathrm{ref}}$ represents the 
prediction from a frozen reference model. The temperature hyperparameter $\beta$, number of diffusion steps $T$, and 
timestep weighting function $\omega (\lambda _{t})$ are used to scale gradients dynamically across sampling steps.

Each term in the DSD loss supports a distinct optimization role: (1) Preference Alignment encourages the model to 
predict noise close to the target noise of the winning image, thereby learning to reproduce 
the generation direction that leads to preferred outcomes. (2) Reference Regularization penalizes the model less when 
the frozen reference already closely matches the winning image. This ensure the model learns where improvement over the 
reference model is possible. (3) Dispreference Repulsion drives the model away from replicating the noise trajectory of 
the losing image, explicitly suppressing undesirable generation behaviors. (4) Reference Normalization preserves the 
baseline behavior of the reference model on dispreferred samples and allows selective suppression of undesirable patterns 
while retaining valid generation capabilities, just like the negative sample rectification mechanism.

Through this design, DSD integrates preference cues directly into the generative denoising process, circumventing 
the need for explicit per-pixel supervision or hand-crafted reward functions. Critically, since the base diffusion 
model remains frozen, the learning is fully captured by the lightweight adapter modules, making the optimization 
both efficient and stable. This setup ensures the model is steered toward generating outputs that more closely 
reflect human-aligned quality, while maintaining subject identity and layout fidelity as per the given instruction.

\begin{figure*}[tbp]
  \centering
  \includegraphics[width=\linewidth]{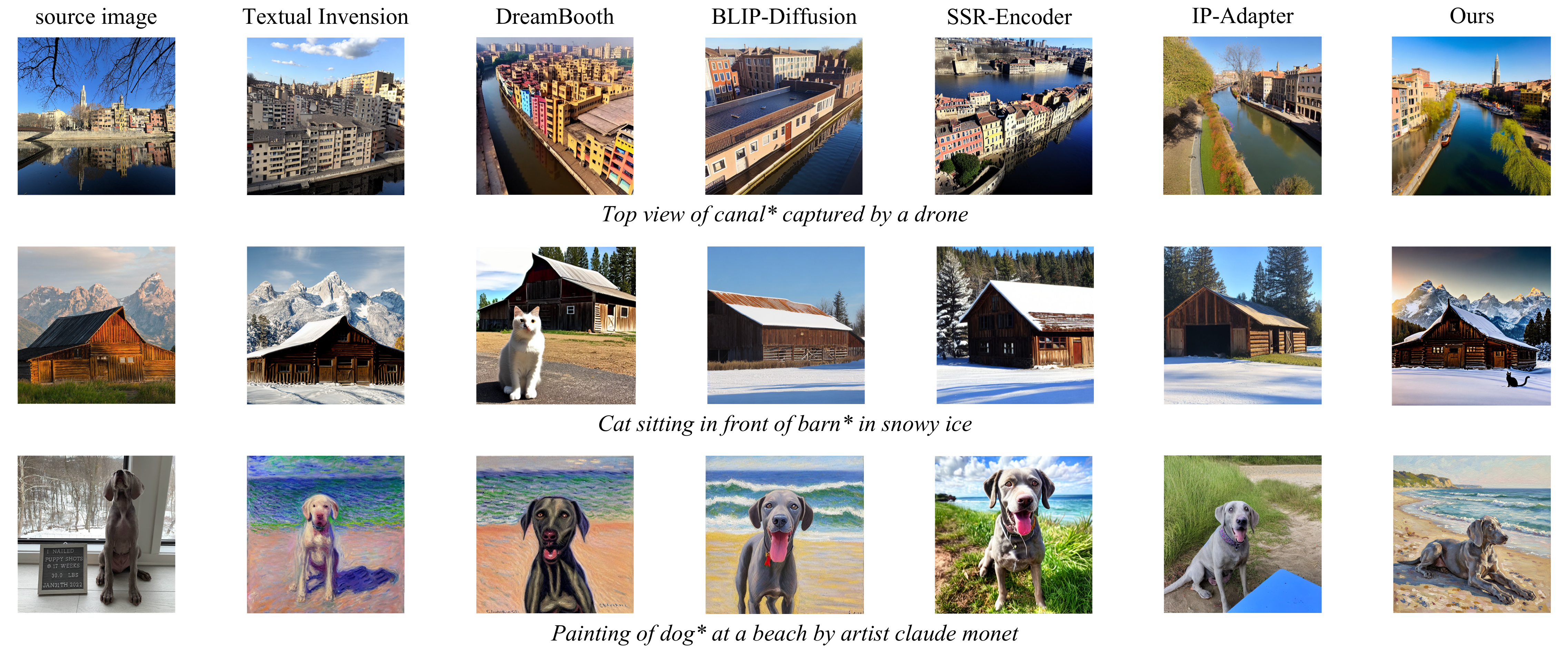}
  \caption{\label{fig:comparison}%
           Quantitative Comparison between TIDE with text-driven image generation methods \cite{gal2022image, ruiz2023dreambooth, li2023blip, zhang2024ssr, ye2023ip}.
           Our model achieves significantly higher subject fidelity and prompt adherence, particularly for complex, long-form textual instruction.}
\end{figure*}

\begin{table*}[t]
\centering
\caption{Quantitative comparisons on Concept101 and DreamBench. DreamBench results are referenced from Subject-Diffusion and SSR-Encoder. 
Concept101 results were tested by ourselves. All methods use SD v1.5 as base model.The best results are \textbf{bolded}, and the second-best results are \underline{underlined}.}
\label{tab:full_results}
\setlength{\tabcolsep}{10pt} 
\begin{tabular}{@{\hspace{1em}}ll|*{3}{S[table-format=1.3]}|*{3}{S[table-format=1.3]}@{\hspace{1em}}}
\toprule
\multirow{2}{*}{Type} & \multirow{2}{*}{Method} & \multicolumn{3}{c|}{Concept101} & \multicolumn{3}{c}{DreamBench} \\
\cmidrule(lr{3pt}){3-5} \cmidrule(lr{3pt}){6-8}
 & & {CLIP-T} & {CLIP-I} & {DINO} & {CLIP-T} & {CLIP-I} & {DINO} \\ 
\midrule
\multirow{3}{*}{Fine-tuning Based} 
& Textual Inversion & 0.280 & 0.786 & 0.670 & 0.255 & 0.780 & 0.569 \\
& DreamBooth & \underline{0.303} & 0.753 & 0.638 & 0.305 & 0.803 & 0.668 \\
& Custom Diffusion & 0.274 & 0.789 & 0.658 & 0.287 & 0.788 & 0.653 \\
\midrule
\multirow{6}{*}{Fine-tuning Free}
& ELITE & 0.295 & 0.763 & 0.611 & 0.298 & 0.775 & 0.605 \\
& BLIP-Diffusion & 0.270 & 0.797 & 0.681 & 0.300 & 0.779 & 0.594 \\
& IP-Adapter & 0.271 & \underline{0.812} & \underline{0.687} & 0.274 & 0.809 & 0.608 \\
& SSR-Encoder & 0.297 & 0.805 & 0.672 & \underline{0.308} & \underline{0.821} & 0.612 \\
& Subject-Diffusion & 0.300 & 0.792 & 0.686 & 0.293 & 0.787 & \textbf{0.711} \\
& TIDE(ours) & \textbf{0.304} & \textbf{0.814} & \textbf{0.690} & \textbf{0.314} & \textbf{0.826} & \underline{0.676} \\
\bottomrule
\end{tabular}
\vspace{2mm}
\end{table*}
\vspace{-0.5cm} 
\section{Experiments}

\subsection{Implementation Details and Evaluation}

Our experimental environment utilized 2  NVIDIA V100 GPUs(32GB). Each training run consisted 
of 50 epochs (40,000 steps) with a batch size of 8, and a learning rate of 1e-5, along with a warmup 
scheduler. Each epoch took approximately 35 minutes to complete. 
Our model is trained on C4DD, with detailed dataset construction described in Sec \ref{sec:data}. In 
order to validate the generation capability in the open domain, we evaluate 
our model not only in Concept101\ \cite{kumari2023multi}, but also follow the DreamBench\ \cite{ruiz2023dreambooth} for 
quantitative and qualitative comparison. DreamBench includes 30 subjects with 25 
instructions while Concept101 includes 101 subjects with different 20 instructions for 
each subject. Although Custom Diffusion provided Concept101 for human preference 
assessment, we keep the automatic evaluation so calculate the CLIP visual similarity 
(CLIP-I) and DINO similarity between the generated images and the target subject 
images as subject alignment, and calculate the CLIP text-image similarity (CLIP-T) 
between the generated images and the given text instruction as text alignment. 

We compare several methods for personalized image generation, including Textual 
Inversion\ \cite{gal2022image}, DreamBooth\ \cite{ruiz2023dreambooth} and Custom Diffusion\ \cite{kumari2023multi}. These methods require test-time fine-tuning 
on given personalized images within specific category. In addition, we compare ELITE\ \cite{wei2023elite}, 
BLIP-Diffusion\ \cite{li2023blip}, IP-adapter\ \cite{ye2023ip}, SSR-Encoder\ \cite{zhang2024ssr} and Subject Diffusion\ \cite{ma2024subject}, all of which trained 
on a large-scale open-domain dataset without test-time fine-tuning. All baseline methods and our proposed 
approach are implemented based on the same Stable Diffusion model (SD v1.5) for fair comparison.

\subsection{Experiment Results}

\textbf{Quantitative comparison}. Table \ref{tab:full_results} presents our quantitative evaluation across 
two benchmarks: Concept101 and DreamBench. We follow DreamBooth to generate 6 images 
for each text prompt provided from DreamBench or Concept101, amounting to a total 4,500 
and 12,120 images for all the subjects. We report the average DINO, CLIP-I and CLIP-T 
scores over all pairs of real and generated images. Our comprehensive evaluation 
across two benchmark datasets demonstrates our method achieves superior scores on all CLIP-I and 
CLIP-T metrics across both test sets, outperforming all eight baseline methods by 
significant margins. In the DreamBench evaluation, while our approach ranks first in 
CLIP-I (with 0.826 score) and CLIP-T (0.314), it attains a competitive second 
position in DINO scoring (0.676), which still exceeds six other comparative methods 
and trails the leader by 3.5\%. This minor variation in a single metric across 
one dataset does not diminish the overall dominance of the method, particularly given 
its consistent top-tier performance in text-alignment (CLIP-T) and 
image-quality (CLIP-I) metrics that are often prioritized for subject-driven 
generation tasks.

Critically, the experiment results on DreamBench show that our model does not merely memorize training 
data. Instead, it successfully handles unseen subject-command combinations, including rare object categories 
and novel pose descriptions, achieving a 6.8\% higher DINO score than the baseline model (IP-Adapter). 
This performance gap underscores our model's robust generalization to unencountered scenarios, a key 
requirement for practical subject-driven generation.

Figure \ref{fig:comparison} presents a qualitative comparison of different SDIG methods  
across diverse prompts. Methods such as BLIP-Diffusion, 
IP-Adapter, SSR-Encoder, Textual Inversion and DreamBooth exhibit noticeable 
degradation in subject fidelity compared to our method. Notably, our method achieves 
subject preservation quality comparable to multi-image DreamBooth fine-tuning, while 
simultaneously maintaining superior text-alignment consistency, in contrast to 
the typical need for 3--5 reference images in DreamBooth.


\begin{figure*}[tbp]
  \centering
  \includegraphics[width=\linewidth]{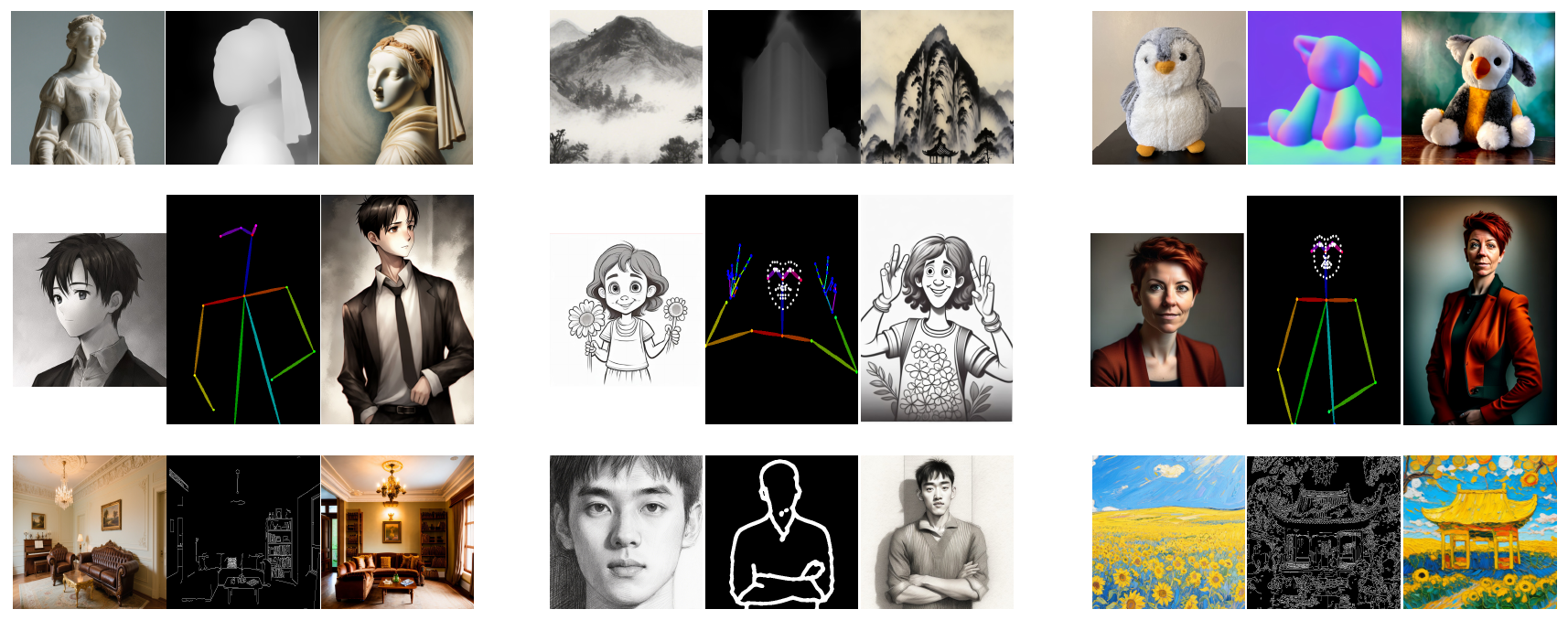}
  \hfill \mbox{}
  \caption{\label{fig:ex8}%
           Visualization of model-generated images: utilizing image prompts under diverse structural conditions.}
\end{figure*}

\subsection{Ablation Studies}

\begin{figure}[h]
  \centering
  \includegraphics[width=\linewidth]{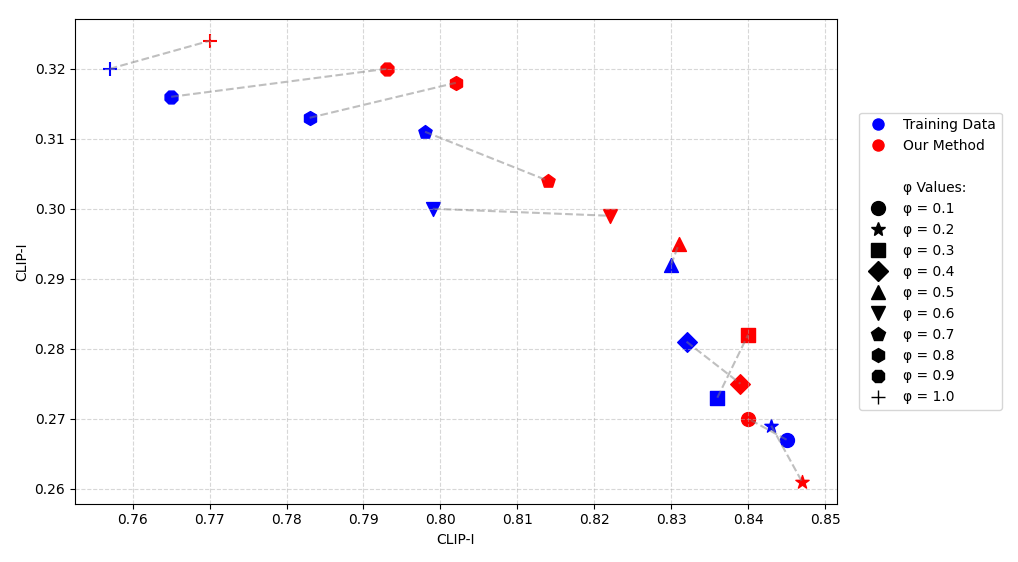}
  \caption{\label{fig:ex7}%
           Ablation study on DSD impact, where data points positioned closer to the 
           top-right corner of the plot indicate superior performance.}
\end{figure}

\textbf{Direct Subject Diffusion Loss Ablation}. While Direct Preference Optimization loss 
was designed for language model alignment, we adapt it to subject-driven image generation through 
DSD loss, a variant that specifically addresses two domain-specific requirements: (1) 
precise appearance feature preservation and (2) instruction-aware subject composition. 
This domain adaptation unlocks nuanced control beyond conventional DPO. 
To further validate the design choices of DSD, we conduct controlled experiments on the C4DD 
dataset, varying the fusion weight $\varphi$ in Eq.\eqref{eq:eq13} that governs text-visual preference 
balance. As shown as Figure~\ref{fig:ex7}, we find $\varphi=0.7$ achieves peak performance 
(0.814 CLIP-I, 0.304 CLIP-T), demonstrating a 3\% improvement over the training set reference 
images. This confirms the necessity of balanced text-visual alignment in subject-driven tasks. 
Across all $\varphi$ configurations, our method demonstrates consistent improvements over the 
original training data, achieving an average 1.9\% enhancement in the composite metric. This marginal 
but statistically significant gain confirms that DSD optimization effectively distills transferable 
knowledge beyond mere dataset replication preference-based contrastive learning. This ablation 
comfirms the role of DSD as a domain-optimized instantiation of DPO for subject-driven image generation, where 
simultaneous control of subject identity and editable attributes is paramount. 

\begin{table}[H]
\centering
\caption{Ablation study of different components}
\label{tab:ablation}
\footnotesize 
\begin{tabular}{@{}lc*{3}{S[table-format=1.3]}@{}}
\toprule
\multirow{2}{*}{Method} & \multirow{2}{*}{Index} & \multicolumn{3}{c}{Evaluation Metrics} \\
\cmidrule(lr){3-5}
 & & {CLIP-T} & {CLIP-I} & {DINO} \\ 
\midrule
Full Model & (a) & \textbf{0.314} & \textbf{0.826} & \textbf{0.676} \\
w/o trainning on C4DD & (b) & 0.274 & 0.809 & 0.608 \\
w/o DSD loss & (c) & 0.275 & 0.820 & 0.665 \\
w/o target supervision & (d) & 0.265 & 0.815 & 0.668 \\
w/o IPM-ICAM & (e) & 0.299 & 0.719 & 0.637 \\
\bottomrule
\end{tabular}
\vspace{-3mm} 
\end{table}

\textbf{Different Components Ablation}. Table \ref{tab:ablation} presents zero-shot evaluation 
results on DreamBench, systematically assessing the impact of individual components in our 
proposed framework. Each ablation setting (b--e) exhibits quantitatively inferior performance 
compared to the full model (a), underscoring the critical role of each component in achieving 
optimal results.

Experiment (b) serves as the baseline model (IP-Adapter), trained solely on OpenImages with self-supervised 
reconstruction objectives. Our evaluation results (a) and (b) demonstrate that integrating 
our C4DD dataset significantly enhances the generative capabilities of the baseline. This confirms 
that the tailored dataset effectively bridges the gap between generic vision-language alignment 
and subject-driven image generation tasks. 

The comparison between (a) and (c) isolates the contribution of the DSD loss function. The 
inclusion of DSD yields substantial gains across all metrics (with the CLIP-T score 
increasing by 3.9\%, the CLIP-I increasing by 0.6\%, and DINO score increasing by 1.9\%). 
This improvement is particularly pronounced in semantic fidelity (CLIP-T), suggesting 
that DSD loss plays a crucial role in enabling the model to effectively capture 
fine-grained semantic nuances.

Experiments (a) vs. (d) highlight the importance of target supervision. Following the 
self-supervised reconstruction experiment setup, using the "winning" images as input images
in experiment (d) leads to a 4.9\% drop in the CLIP-T metric. This degradation arises 
because the small scale of C4DD is insufficient to support robust self-supervised learning 
without explicit guidance.

The results of (a) and (e) emphasize the criticality of the IPM-ICAM feature fusion mechanism. 
Removing this component results in a precipitous decline in CLIP-I (-10.7\%), indicating a 
loss of subject fidelity. This demonstrates that IPM-ICAM effectively aggregates multimodal 
features to preserve subject characteristics during generation.

\begin{figure}[t]
  \centering
  \includegraphics[width=\linewidth]{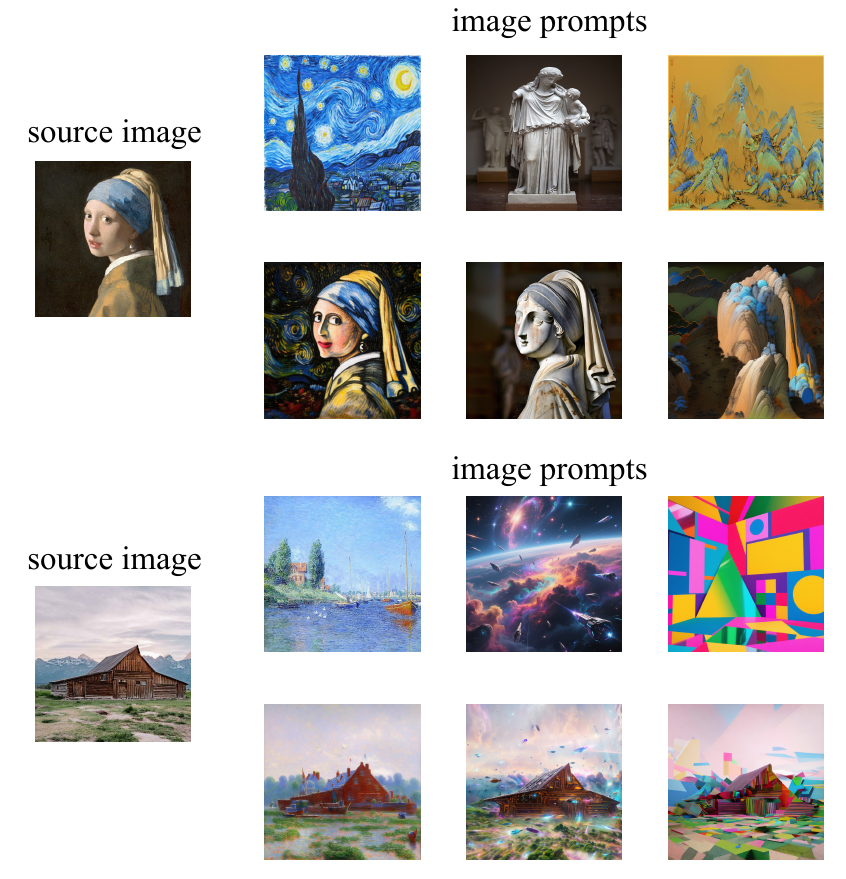}
  \caption{\label{fig:ex9}%
           Visualization of image-to-image by our model.}
\end{figure}

\subsection{More Applications}
Although the proposed TIDE is designed to achieve the generation with image prompts, 
its robust generalization capabilities allow for a broader range of applications. 
TIDE is not only efficient in subject-driven image generation but also compatible with existing 
controllable tools and text prompts. In this part, we show more results that our
model can generate.

\textbf{Structural Condition}. A pivotal advantage of text-to-image diffusion models lies 
in their capacity for structural-conditioned generation. Crucially, the adapter 
architecture of TIDE preserves the native compatibility of base model with existing control 
mechanisms (e.g., ControlNet\ \cite{zhang2023adding}), enabling multi-condition synthesis.
As demonstrated in Figure \ref{fig:ex8}, TIDE successfully processes diverse control 
modalities, including depth map, normal map, human pose skeleton, canny edge and 
free-form scribble. Notably, this functionality requires no additional fine-tuning 
and the parameter-efficient design of the adapter
ensures seamless interoperability with control tools while avoiding catastrophic 
interference. The simultaneous satisfaction of structural constraints and subject 
preservation demonstrates the capability of TIDE for multi-condition generation without 
control-specific tuning.

\textbf{Image-to-Image Generation}. TIDE naturally extends to image-to-image style 
transfer by leveraging its core subject-preservation architecture. Given a content 
image and style reference, the framework seamlessly transfers artistic attributes 
(e.g., brushstrokes, color palettes) while rigorously maintaining the structural 
integrity of the original subject. As illustrated in Figure \ref{fig:ex9}, this enables 
convincing transformations across diverse media (from watercolor renditions to 
synthetic art styles) without distorting key subject characteristics. Honed during 
preference optimization, the adapter innately understands subject-style relationships, 
enabling it to intelligently balance stylistic adaptation with identity preservation. 
This capability outperforms conventional style transfer tools, which often compromise subject fidelity.

\begin{figure}[H]
  \centering
  \includegraphics[width=\linewidth]{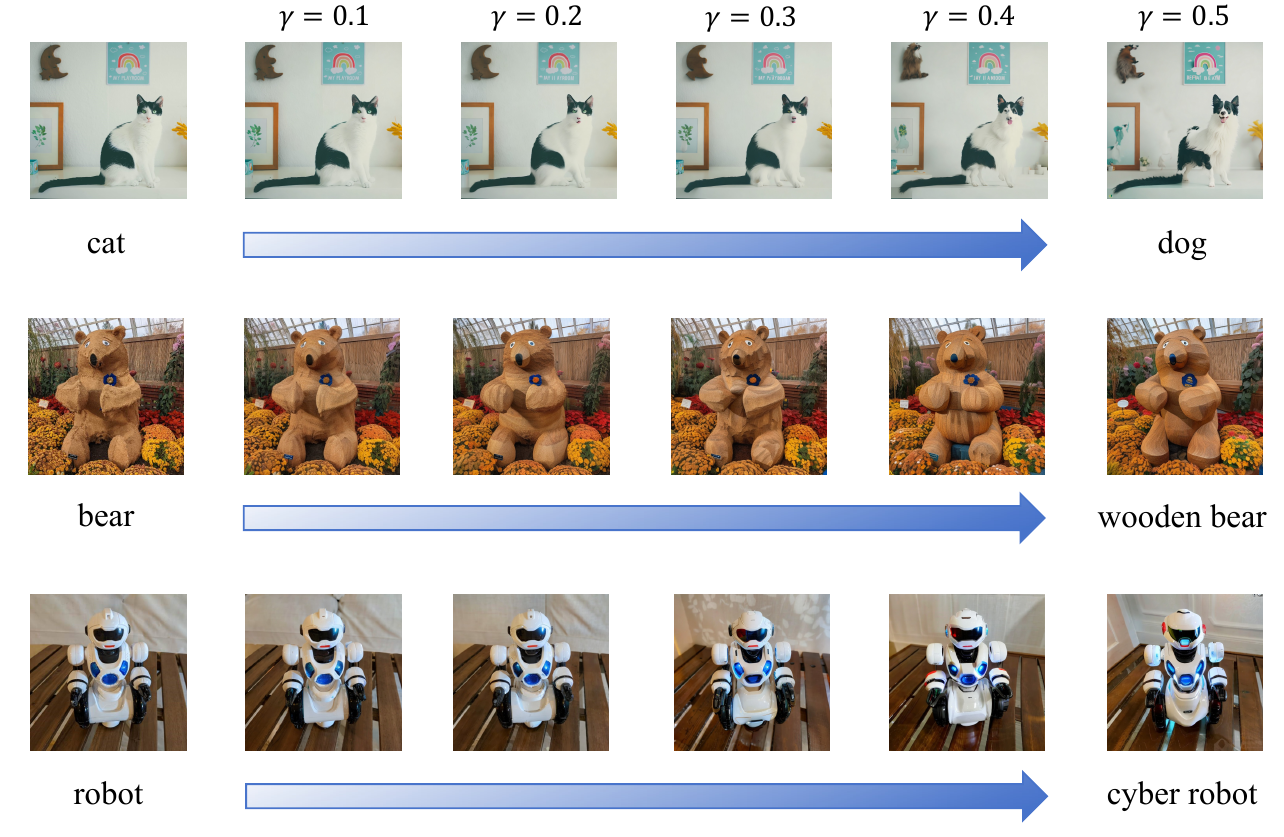}
  \caption{\label{fig:ex10}%
           Text-image interpolation.}
\end{figure}

\textbf{Latent Space Interpolation}. As formalized in Eq.\eqref{eq:eq11}, TIDE governs cross-modal interpolation 
through the tunable parameter $\gamma$, which dynamically weights the contributions of text and image embeddings 
to precisely control the balance between text and image references during attribute transitions. As demonstrated in Figure \ref{fig:ex10}, smooth interpolations
across subject, material and style attributes are achieved within a $\gamma$ range of 0.1 to 0.5, while 
maintaining exceptional background consistency and subject fidelity. This smoothness ensures artistically 
viable transitions without perceptual discontinuities. When $\gamma$ exceeds 0.5, the model's attention to visual references 
diminishes, causing a gradual degradation into conventional text-to-image generation as textual 
guidance becomes disproportionately dominant. This behavior is not a failure mode but rather an 
expected outcome of our training strategy, which prioritizes structural alignment through carefully calibrated $\gamma$ values 
to ensure robust subject preservation while still allowing flexible attribute manipulation.


\section{Conclusion}
TIDE establishes a new framework for subject-driven image generation by resolving the 
fundamental preservation-compliance trade-off through target-supervised learning and 
preference optimization. Our framework eliminates test-time fine-tuning via a lightweight 
multimodal adapter that fuses visual-textual features while preserving base model integrity.
Methodologically, this represents a paradigm shift from reconstruction-based self-supervision 
to instruction-target aligned learning, enabling unprecedented precision in subject-instruction coordination.

Extensive validation demonstrates the superiority of TIDE in maintaining subject fidelity ($\uparrow 6.8\%$ DINO score over the baseline model) 
and instruction compliance ($>31\%$ CLIP-T score), with zero-shot adaptability to structural 
editing, image-to-image translation, and interpolation tasks. These capabilities position TIDE 
as a foundational tool for creative industries requiring reliable subject manipulation, 
from personalized advertising to educational content generation. This work bridges the gap between 
precise subject control and flexible instruction following, opening avenues for instruction-aware 
generative systems.

\section{Limitation}

Although our method enables zero-shot generation for arbitrary open-domain subjects, 
it still has certain limitations. The framework inherits the text comprehension constraints of CLIP, 
and notably, existing methods have not yet focused on addressing the fragmented or complex paragraph-length instructions which exceed typical token 
thresholds. Additionally, handling minority languages may trigger subject-instruction 
misalignment due to insufficient multilingual training data, potentially leading to hallucinated 
outputs in non-dominant language scenarios. In the future, we plan to integrate additional techniques to overcome these 
limitations while maintaining the efficiency benefits of our current paradigm.

\bibliographystyle{ACM-Reference-Format}
\bibliography{TIDE_sub.bib}       


\begin{thebibliography}{47}


\ifx \showCODEN    \undefined \def \showCODEN     #1{\unskip}     \fi
\ifx \showISBNx    \undefined \def \showISBNx     #1{\unskip}     \fi
\ifx \showISBNxiii \undefined \def \showISBNxiii  #1{\unskip}     \fi
\ifx \showISSN     \undefined \def \showISSN      #1{\unskip}     \fi
\ifx \showLCCN     \undefined \def \showLCCN      #1{\unskip}     \fi
\ifx \shownote     \undefined \def \shownote      #1{#1}          \fi
\ifx \showarticletitle \undefined \def \showarticletitle #1{#1}   \fi
\ifx \showURL      \undefined \def \showURL       {\relax}        \fi
\providecommand\bibfield[2]{#2}
\providecommand\bibinfo[2]{#2}
\providecommand\natexlab[1]{#1}
\providecommand\showeprint[2][]{arXiv:#2}

\bibitem[Achiam et~al\mbox{.}(2023)]%
        {achiam2023gpt}
\bibfield{author}{\bibinfo{person}{Josh Achiam}, \bibinfo{person}{Steven
  Adler}, \bibinfo{person}{Sandhini Agarwal}, \bibinfo{person}{Lama Ahmad},
  \bibinfo{person}{Ilge Akkaya}, \bibinfo{person}{Florencia~Leoni Aleman},
  \bibinfo{person}{Diogo Almeida}, \bibinfo{person}{Janko Altenschmidt},
  \bibinfo{person}{Sam Altman}, \bibinfo{person}{Shyamal Anadkat},
  {et~al\mbox{.}}} \bibinfo{year}{2023}\natexlab{}.
\newblock \showarticletitle{Gpt-4 technical report}.
\newblock \bibinfo{journal}{\emph{arXiv preprint arXiv:2303.08774}}
  (\bibinfo{year}{2023}).
\newblock


\bibitem[Ahn et~al\mbox{.}(2024)]%
        {ahn2024self}
\bibfield{author}{\bibinfo{person}{Donghoon Ahn}, \bibinfo{person}{Hyoungwon
  Cho}, \bibinfo{person}{Jaewon Min}, \bibinfo{person}{Wooseok Jang},
  \bibinfo{person}{Jungwoo Kim}, \bibinfo{person}{SeonHwa Kim},
  \bibinfo{person}{Hyun~Hee Park}, \bibinfo{person}{Kyong~Hwan Jin}, {and}
  \bibinfo{person}{Seungryong Kim}.} \bibinfo{year}{2024}\natexlab{}.
\newblock \showarticletitle{Self-rectifying diffusion sampling with
  perturbed-attention guidance}. In \bibinfo{booktitle}{\emph{European
  Conference on Computer Vision}}. Springer, \bibinfo{pages}{1--17}.
\newblock


\bibitem[Avrahami et~al\mbox{.}(2023)]%
        {avrahami2023break}
\bibfield{author}{\bibinfo{person}{Omri Avrahami}, \bibinfo{person}{Kfir
  Aberman}, \bibinfo{person}{Ohad Fried}, \bibinfo{person}{Daniel Cohen-Or},
  {and} \bibinfo{person}{Dani Lischinski}.} \bibinfo{year}{2023}\natexlab{}.
\newblock \showarticletitle{Break-a-scene: Extracting multiple concepts from a
  single image}. In \bibinfo{booktitle}{\emph{SIGGRAPH Asia 2023 Conference
  Papers}}. \bibinfo{pages}{1--12}.
\newblock


\bibitem[Bai et~al\mbox{.}(2022)]%
        {bai2022constitutional}
\bibfield{author}{\bibinfo{person}{Yuntao Bai}, \bibinfo{person}{Saurav
  Kadavath}, \bibinfo{person}{Sandipan Kundu}, \bibinfo{person}{Amanda Askell},
  \bibinfo{person}{Jackson Kernion}, \bibinfo{person}{Andy Jones},
  \bibinfo{person}{Anna Chen}, \bibinfo{person}{Anna Goldie},
  \bibinfo{person}{Azalia Mirhoseini}, \bibinfo{person}{Cameron McKinnon},
  {et~al\mbox{.}}} \bibinfo{year}{2022}\natexlab{}.
\newblock \showarticletitle{Constitutional ai: Harmlessness from ai feedback}.
\newblock \bibinfo{journal}{\emph{arXiv preprint arXiv:2212.08073}}
  (\bibinfo{year}{2022}).
\newblock


\bibitem[Chefer et~al\mbox{.}(2024)]%
        {chefer2024hidden}
\bibfield{author}{\bibinfo{person}{Hila Chefer}, \bibinfo{person}{Oran Lang},
  \bibinfo{person}{Mor Geva}, \bibinfo{person}{Volodymyr Polosukhin},
  \bibinfo{person}{Assaf Shocher}, \bibinfo{person}{Michal Irani},
  \bibinfo{person}{Inbar Mosseri}, {and} \bibinfo{person}{Lior Wolf}.}
  \bibinfo{year}{2024}\natexlab{}.
\newblock \showarticletitle{The Hidden Language of Diffusion Models}. In
  \bibinfo{booktitle}{\emph{12th International Conference on Learning
  Representations, ICLR 2024}}.
\newblock


\bibitem[Dhariwal and Nichol(2021)]%
        {dhariwal2021diffusion}
\bibfield{author}{\bibinfo{person}{Prafulla Dhariwal} {and}
  \bibinfo{person}{Alexander Nichol}.} \bibinfo{year}{2021}\natexlab{}.
\newblock \showarticletitle{Diffusion models beat gans on image synthesis}.
\newblock \bibinfo{journal}{\emph{Advances in neural information processing
  systems}}  \bibinfo{volume}{34} (\bibinfo{year}{2021}),
  \bibinfo{pages}{8780--8794}.
\newblock


\bibitem[Ding et~al\mbox{.}(2024)]%
        {ding2024freecustom}
\bibfield{author}{\bibinfo{person}{Ganggui Ding}, \bibinfo{person}{Canyu Zhao},
  \bibinfo{person}{Wen Wang}, \bibinfo{person}{Zhen Yang},
  \bibinfo{person}{Zide Liu}, \bibinfo{person}{Hao Chen}, {and}
  \bibinfo{person}{Chunhua Shen}.} \bibinfo{year}{2024}\natexlab{}.
\newblock \showarticletitle{Freecustom: Tuning-free customized image generation
  for multi-concept composition}. In \bibinfo{booktitle}{\emph{Proceedings of
  the IEEE/CVF Conference on Computer Vision and Pattern Recognition}}.
  \bibinfo{pages}{9089--9098}.
\newblock


\bibitem[Dong et~al\mbox{.}(2023)]%
        {dong2023abilities}
\bibfield{author}{\bibinfo{person}{Guanting Dong}, \bibinfo{person}{Hongyi
  Yuan}, \bibinfo{person}{Keming Lu}, \bibinfo{person}{Chengpeng Li},
  \bibinfo{person}{Mingfeng Xue}, \bibinfo{person}{Dayiheng Liu},
  \bibinfo{person}{Wei Wang}, \bibinfo{person}{Zheng Yuan},
  \bibinfo{person}{Chang Zhou}, {and} \bibinfo{person}{Jingren Zhou}.}
  \bibinfo{year}{2023}\natexlab{}.
\newblock \showarticletitle{How abilities in large language models are affected
  by supervised fine-tuning data composition}.
\newblock \bibinfo{journal}{\emph{arXiv preprint arXiv:2310.05492}}
  (\bibinfo{year}{2023}).
\newblock


\bibitem[Dubois et~al\mbox{.}(2023)]%
        {dubois2023alpacafarm}
\bibfield{author}{\bibinfo{person}{Yann Dubois}, \bibinfo{person}{Chen~Xuechen
  Li}, \bibinfo{person}{Rohan Taori}, \bibinfo{person}{Tianyi Zhang},
  \bibinfo{person}{Ishaan Gulrajani}, \bibinfo{person}{Jimmy Ba},
  \bibinfo{person}{Carlos Guestrin}, \bibinfo{person}{Percy~S Liang}, {and}
  \bibinfo{person}{Tatsunori~B Hashimoto}.} \bibinfo{year}{2023}\natexlab{}.
\newblock \showarticletitle{Alpacafarm: A simulation framework for methods that
  learn from human feedback}.
\newblock \bibinfo{journal}{\emph{Advances in Neural Information Processing
  Systems}}  \bibinfo{volume}{36} (\bibinfo{year}{2023}),
  \bibinfo{pages}{30039--30069}.
\newblock


\bibitem[Gal et~al\mbox{.}(2022)]%
        {gal2022image}
\bibfield{author}{\bibinfo{person}{Rinon Gal}, \bibinfo{person}{Yuval Alaluf},
  \bibinfo{person}{Yuval Atzmon}, \bibinfo{person}{Or Patashnik},
  \bibinfo{person}{Amit~H Bermano}, \bibinfo{person}{Gal Chechik}, {and}
  \bibinfo{person}{Daniel Cohen-Or}.} \bibinfo{year}{2022}\natexlab{}.
\newblock \showarticletitle{An image is worth one word: Personalizing
  text-to-image generation using textual inversion}.
\newblock \bibinfo{journal}{\emph{arXiv preprint arXiv:2208.01618}}
  (\bibinfo{year}{2022}).
\newblock


\bibitem[Gal et~al\mbox{.}(2023)]%
        {gal2023encoder}
\bibfield{author}{\bibinfo{person}{Rinon Gal}, \bibinfo{person}{Moab Arar},
  \bibinfo{person}{Yuval Atzmon}, \bibinfo{person}{Amit~H Bermano},
  \bibinfo{person}{Gal Chechik}, {and} \bibinfo{person}{Daniel Cohen-Or}.}
  \bibinfo{year}{2023}\natexlab{}.
\newblock \showarticletitle{Encoder-based domain tuning for fast
  personalization of text-to-image models}.
\newblock \bibinfo{journal}{\emph{ACM Transactions on Graphics (TOG)}}
  \bibinfo{volume}{42}, \bibinfo{number}{4} (\bibinfo{year}{2023}),
  \bibinfo{pages}{1--13}.
\newblock


\bibitem[Han et~al\mbox{.}(2023)]%
        {han2023svdiff}
\bibfield{author}{\bibinfo{person}{Ligong Han}, \bibinfo{person}{Yinxiao Li},
  \bibinfo{person}{Han Zhang}, \bibinfo{person}{Peyman Milanfar},
  \bibinfo{person}{Dimitris Metaxas}, {and} \bibinfo{person}{Feng Yang}.}
  \bibinfo{year}{2023}\natexlab{}.
\newblock \showarticletitle{Svdiff: Compact parameter space for diffusion
  fine-tuning}. In \bibinfo{booktitle}{\emph{Proceedings of the IEEE/CVF
  International Conference on Computer Vision}}. \bibinfo{pages}{7323--7334}.
\newblock


\bibitem[Hyung et~al\mbox{.}(2024)]%
        {hyung2024magicapture}
\bibfield{author}{\bibinfo{person}{Junha Hyung}, \bibinfo{person}{Jaeyo Shin},
  {and} \bibinfo{person}{Jaegul Choo}.} \bibinfo{year}{2024}\natexlab{}.
\newblock \showarticletitle{Magicapture: High-resolution multi-concept portrait
  customization}. In \bibinfo{booktitle}{\emph{Proceedings of the AAAI
  Conference on Artificial Intelligence}}, Vol.~\bibinfo{volume}{38}.
  \bibinfo{pages}{2445--2453}.
\newblock


\bibitem[Imrey(2005)]%
        {imrey2005b}
\bibfield{author}{\bibinfo{person}{Peter~B Imrey}.}
  \bibinfo{year}{2005}\natexlab{}.
\newblock \showarticletitle{B radley--T erry Model}.
\newblock \bibinfo{journal}{\emph{Encyclopedia of Biostatistics}}
  \bibinfo{volume}{1} (\bibinfo{year}{2005}).
\newblock


\bibitem[Kingma et~al\mbox{.}(2021)]%
        {kingma2021variational}
\bibfield{author}{\bibinfo{person}{Diederik Kingma}, \bibinfo{person}{Tim
  Salimans}, \bibinfo{person}{Ben Poole}, {and} \bibinfo{person}{Jonathan Ho}.}
  \bibinfo{year}{2021}\natexlab{}.
\newblock \showarticletitle{Variational diffusion models}.
\newblock \bibinfo{journal}{\emph{Advances in neural information processing
  systems}}  \bibinfo{volume}{34} (\bibinfo{year}{2021}),
  \bibinfo{pages}{21696--21707}.
\newblock


\bibitem[Krishnaiah et~al\mbox{.}(2024)]%
        {krishnaiah2024harmonizing}
\bibfield{author}{\bibinfo{person}{N Krishnaiah}, \bibinfo{person}{Vastrala
  Vaishnavi}, \bibinfo{person}{Ch Thanmai}, \bibinfo{person}{D Pranathi},
  \bibinfo{person}{G Sunaina}, \bibinfo{person}{K~Gayathri Bhavya}, {and}
  \bibinfo{person}{M Aditi}.} \bibinfo{year}{2024}\natexlab{}.
\newblock \showarticletitle{Harmonizing Offline Reinforcement Learning with
  Language Models Analysis of Human Responses}.
\newblock \bibinfo{journal}{\emph{learning}} \bibinfo{volume}{24},
  \bibinfo{number}{2} (\bibinfo{year}{2024}).
\newblock


\bibitem[Kumari et~al\mbox{.}(2023)]%
        {kumari2023multi}
\bibfield{author}{\bibinfo{person}{Nupur Kumari}, \bibinfo{person}{Bingliang
  Zhang}, \bibinfo{person}{Richard Zhang}, \bibinfo{person}{Eli Shechtman},
  {and} \bibinfo{person}{Jun-Yan Zhu}.} \bibinfo{year}{2023}\natexlab{}.
\newblock \showarticletitle{Multi-concept customization of text-to-image
  diffusion}. In \bibinfo{booktitle}{\emph{Proceedings of the IEEE/CVF
  conference on computer vision and pattern recognition}}.
  \bibinfo{pages}{1931--1941}.
\newblock


\bibitem[Kuznetsova et~al\mbox{.}(2020)]%
        {kuznetsova2020open}
\bibfield{author}{\bibinfo{person}{Alina Kuznetsova}, \bibinfo{person}{Hassan
  Rom}, \bibinfo{person}{Neil Alldrin}, \bibinfo{person}{Jasper Uijlings},
  \bibinfo{person}{Ivan Krasin}, \bibinfo{person}{Jordi Pont-Tuset},
  \bibinfo{person}{Shahab Kamali}, \bibinfo{person}{Stefan Popov},
  \bibinfo{person}{Matteo Malloci}, \bibinfo{person}{Alexander Kolesnikov},
  {et~al\mbox{.}}} \bibinfo{year}{2020}\natexlab{}.
\newblock \showarticletitle{The open images dataset v4: Unified image
  classification, object detection, and visual relationship detection at
  scale}.
\newblock \bibinfo{journal}{\emph{International journal of computer vision}}
  \bibinfo{volume}{128}, \bibinfo{number}{7} (\bibinfo{year}{2020}),
  \bibinfo{pages}{1956--1981}.
\newblock


\bibitem[Li et~al\mbox{.}(2023)]%
        {li2023blip}
\bibfield{author}{\bibinfo{person}{Dongxu Li}, \bibinfo{person}{Junnan Li},
  {and} \bibinfo{person}{Steven Hoi}.} \bibinfo{year}{2023}\natexlab{}.
\newblock \showarticletitle{Blip-diffusion: Pre-trained subject representation
  for controllable text-to-image generation and editing}.
\newblock \bibinfo{journal}{\emph{Advances in Neural Information Processing
  Systems}}  \bibinfo{volume}{36} (\bibinfo{year}{2023}),
  \bibinfo{pages}{30146--30166}.
\newblock


\bibitem[Lin et~al\mbox{.}(2024)]%
        {lin2024accdiffusion}
\bibfield{author}{\bibinfo{person}{Zhihang Lin}, \bibinfo{person}{Mingbao Lin},
  \bibinfo{person}{Meng Zhao}, {and} \bibinfo{person}{Rongrong Ji}.}
  \bibinfo{year}{2024}\natexlab{}.
\newblock \showarticletitle{Accdiffusion: An accurate method for
  higher-resolution image generation}. In \bibinfo{booktitle}{\emph{European
  Conference on Computer Vision}}. Springer, \bibinfo{pages}{38--53}.
\newblock


\bibitem[Liu et~al\mbox{.}(2023a)]%
        {liu2023cones}
\bibfield{author}{\bibinfo{person}{Zhiheng Liu}, \bibinfo{person}{Ruili Feng},
  \bibinfo{person}{Kai Zhu}, \bibinfo{person}{Yifei Zhang},
  \bibinfo{person}{Kecheng Zheng}, \bibinfo{person}{Yu Liu},
  \bibinfo{person}{Deli Zhao}, \bibinfo{person}{Jingren Zhou}, {and}
  \bibinfo{person}{Yang Cao}.} \bibinfo{year}{2023}\natexlab{a}.
\newblock \showarticletitle{Cones: Concept neurons in diffusion models for
  customized generation}.
\newblock \bibinfo{journal}{\emph{arXiv preprint arXiv:2303.05125}}
  (\bibinfo{year}{2023}).
\newblock


\bibitem[Liu et~al\mbox{.}(2023b)]%
        {liu2023cones2}
\bibfield{author}{\bibinfo{person}{Zhiheng Liu}, \bibinfo{person}{Yifei Zhang},
  \bibinfo{person}{Shen Yujun}, \bibinfo{person}{Zheng Kecheng},
  \bibinfo{person}{Kai Zhu}, \bibinfo{person}{Ruili Feng}, \bibinfo{person}{Liu
  Yu}, \bibinfo{person}{Deli Zhao}, \bibinfo{person}{Jingren Zhou}, {and}
  \bibinfo{person}{Cao Yang}.} \bibinfo{year}{2023}\natexlab{b}.
\newblock \showarticletitle{Cones 2: Customizable Image Synthesis with Multiple
  Subjects}.
\newblock \bibinfo{journal}{\emph{arXiv preprint arXiv:2305.19327}}
  (\bibinfo{year}{2023}).
\newblock


\bibitem[Ma et~al\mbox{.}(2024)]%
        {ma2024subject}
\bibfield{author}{\bibinfo{person}{Jian Ma}, \bibinfo{person}{Junhao Liang},
  \bibinfo{person}{Chen Chen}, {and} \bibinfo{person}{Haonan Lu}.}
  \bibinfo{year}{2024}\natexlab{}.
\newblock \showarticletitle{Subject-diffusion: Open domain personalized
  text-to-image generation without test-time fine-tuning}. In
  \bibinfo{booktitle}{\emph{ACM SIGGRAPH 2024 Conference Papers}}.
  \bibinfo{pages}{1--12}.
\newblock


\bibitem[Ma et~al\mbox{.}(2023)]%
        {ma2023unified}
\bibfield{author}{\bibinfo{person}{Yiyang Ma}, \bibinfo{person}{Huan Yang},
  \bibinfo{person}{Wenjing Wang}, \bibinfo{person}{Jianlong Fu}, {and}
  \bibinfo{person}{Jiaying Liu}.} \bibinfo{year}{2023}\natexlab{}.
\newblock \showarticletitle{Unified multi-modal latent diffusion for joint
  subject and text conditional image generation}.
\newblock \bibinfo{journal}{\emph{arXiv preprint arXiv:2303.09319}}
  (\bibinfo{year}{2023}).
\newblock


\bibitem[Mnih et~al\mbox{.}(2016)]%
        {mnih2016asynchronous}
\bibfield{author}{\bibinfo{person}{Volodymyr Mnih},
  \bibinfo{person}{Adria~Puigdomenech Badia}, \bibinfo{person}{Mehdi Mirza},
  \bibinfo{person}{Alex Graves}, \bibinfo{person}{Timothy Lillicrap},
  \bibinfo{person}{Tim Harley}, \bibinfo{person}{David Silver}, {and}
  \bibinfo{person}{Koray Kavukcuoglu}.} \bibinfo{year}{2016}\natexlab{}.
\newblock \showarticletitle{Asynchronous methods for deep reinforcement
  learning}. In \bibinfo{booktitle}{\emph{International conference on machine
  learning}}. PmLR, \bibinfo{pages}{1928--1937}.
\newblock


\bibitem[Podell et~al\mbox{.}(2023)]%
        {podell2023sdxl}
\bibfield{author}{\bibinfo{person}{Dustin Podell}, \bibinfo{person}{Zion
  English}, \bibinfo{person}{Kyle Lacey}, \bibinfo{person}{Andreas Blattmann},
  \bibinfo{person}{Tim Dockhorn}, \bibinfo{person}{Jonas M{\"u}ller},
  \bibinfo{person}{Joe Penna}, {and} \bibinfo{person}{Robin Rombach}.}
  \bibinfo{year}{2023}\natexlab{}.
\newblock \showarticletitle{Sdxl: Improving latent diffusion models for
  high-resolution image synthesis}.
\newblock \bibinfo{journal}{\emph{arXiv preprint arXiv:2307.01952}}
  (\bibinfo{year}{2023}).
\newblock


\bibitem[Rafailov et~al\mbox{.}(2023)]%
        {rafailov2023direct}
\bibfield{author}{\bibinfo{person}{Rafael Rafailov}, \bibinfo{person}{Archit
  Sharma}, \bibinfo{person}{Eric Mitchell}, \bibinfo{person}{Christopher~D
  Manning}, \bibinfo{person}{Stefano Ermon}, {and} \bibinfo{person}{Chelsea
  Finn}.} \bibinfo{year}{2023}\natexlab{}.
\newblock \showarticletitle{Direct preference optimization: Your language model
  is secretly a reward model}.
\newblock \bibinfo{journal}{\emph{Advances in Neural Information Processing
  Systems}}  \bibinfo{volume}{36} (\bibinfo{year}{2023}),
  \bibinfo{pages}{53728--53741}.
\newblock


\bibitem[Raffel et~al\mbox{.}(2020)]%
        {raffel2020exploring}
\bibfield{author}{\bibinfo{person}{Colin Raffel}, \bibinfo{person}{Noam
  Shazeer}, \bibinfo{person}{Adam Roberts}, \bibinfo{person}{Katherine Lee},
  \bibinfo{person}{Sharan Narang}, \bibinfo{person}{Michael Matena},
  \bibinfo{person}{Yanqi Zhou}, \bibinfo{person}{Wei Li}, {and}
  \bibinfo{person}{Peter~J Liu}.} \bibinfo{year}{2020}\natexlab{}.
\newblock \showarticletitle{Exploring the limits of transfer learning with a
  unified text-to-text transformer}.
\newblock \bibinfo{journal}{\emph{Journal of machine learning research}}
  \bibinfo{volume}{21}, \bibinfo{number}{140} (\bibinfo{year}{2020}),
  \bibinfo{pages}{1--67}.
\newblock


\bibitem[Ramesh et~al\mbox{.}(2022)]%
        {ramesh2022hierarchical}
\bibfield{author}{\bibinfo{person}{Aditya Ramesh}, \bibinfo{person}{Prafulla
  Dhariwal}, \bibinfo{person}{Alex Nichol}, \bibinfo{person}{Casey Chu}, {and}
  \bibinfo{person}{Mark Chen}.} \bibinfo{year}{2022}\natexlab{}.
\newblock \showarticletitle{Hierarchical text-conditional image generation with
  clip latents}.
\newblock \bibinfo{journal}{\emph{arXiv preprint arXiv:2204.06125}}
  \bibinfo{volume}{1}, \bibinfo{number}{2} (\bibinfo{year}{2022}),
  \bibinfo{pages}{3}.
\newblock


\bibitem[Ramesh et~al\mbox{.}(2021)]%
        {ramesh2021zero}
\bibfield{author}{\bibinfo{person}{Aditya Ramesh}, \bibinfo{person}{Mikhail
  Pavlov}, \bibinfo{person}{Gabriel Goh}, \bibinfo{person}{Scott Gray},
  \bibinfo{person}{Chelsea Voss}, \bibinfo{person}{Alec Radford},
  \bibinfo{person}{Mark Chen}, {and} \bibinfo{person}{Ilya Sutskever}.}
  \bibinfo{year}{2021}\natexlab{}.
\newblock \showarticletitle{Zero-shot text-to-image generation}. In
  \bibinfo{booktitle}{\emph{International conference on machine learning}}.
  Pmlr, \bibinfo{pages}{8821--8831}.
\newblock


\bibitem[Rombach et~al\mbox{.}(2022)]%
        {rombach2022high}
\bibfield{author}{\bibinfo{person}{Robin Rombach}, \bibinfo{person}{Andreas
  Blattmann}, \bibinfo{person}{Dominik Lorenz}, \bibinfo{person}{Patrick
  Esser}, {and} \bibinfo{person}{Bj{\"o}rn Ommer}.}
  \bibinfo{year}{2022}\natexlab{}.
\newblock \showarticletitle{High-resolution image synthesis with latent
  diffusion models}. In \bibinfo{booktitle}{\emph{Proceedings of the IEEE/CVF
  conference on computer vision and pattern recognition}}.
  \bibinfo{pages}{10684--10695}.
\newblock


\bibitem[Ruiz et~al\mbox{.}(2023)]%
        {ruiz2023dreambooth}
\bibfield{author}{\bibinfo{person}{Nataniel Ruiz}, \bibinfo{person}{Yuanzhen
  Li}, \bibinfo{person}{Varun Jampani}, \bibinfo{person}{Yael Pritch},
  \bibinfo{person}{Michael Rubinstein}, {and} \bibinfo{person}{Kfir Aberman}.}
  \bibinfo{year}{2023}\natexlab{}.
\newblock \showarticletitle{Dreambooth: Fine tuning text-to-image diffusion
  models for subject-driven generation}. In
  \bibinfo{booktitle}{\emph{Proceedings of the IEEE/CVF conference on computer
  vision and pattern recognition}}. \bibinfo{pages}{22500--22510}.
\newblock


\bibitem[Saharia et~al\mbox{.}(2022)]%
        {saharia2022photorealistic}
\bibfield{author}{\bibinfo{person}{Chitwan Saharia}, \bibinfo{person}{William
  Chan}, \bibinfo{person}{Saurabh Saxena}, \bibinfo{person}{Lala Li},
  \bibinfo{person}{Jay Whang}, \bibinfo{person}{Emily~L Denton},
  \bibinfo{person}{Kamyar Ghasemipour}, \bibinfo{person}{Raphael
  Gontijo~Lopes}, \bibinfo{person}{Burcu Karagol~Ayan}, \bibinfo{person}{Tim
  Salimans}, {et~al\mbox{.}}} \bibinfo{year}{2022}\natexlab{}.
\newblock \showarticletitle{Photorealistic text-to-image diffusion models with
  deep language understanding}.
\newblock \bibinfo{journal}{\emph{Advances in neural information processing
  systems}}  \bibinfo{volume}{35} (\bibinfo{year}{2022}),
  \bibinfo{pages}{36479--36494}.
\newblock


\bibitem[Schuhmann et~al\mbox{.}(2022)]%
        {schuhmann2022laion}
\bibfield{author}{\bibinfo{person}{Christoph Schuhmann},
  \bibinfo{person}{Romain Beaumont}, \bibinfo{person}{Richard Vencu},
  \bibinfo{person}{Cade Gordon}, \bibinfo{person}{Ross Wightman},
  \bibinfo{person}{Mehdi Cherti}, \bibinfo{person}{Theo Coombes},
  \bibinfo{person}{Aarush Katta}, \bibinfo{person}{Clayton Mullis},
  \bibinfo{person}{Mitchell Wortsman}, {et~al\mbox{.}}}
  \bibinfo{year}{2022}\natexlab{}.
\newblock \showarticletitle{Laion-5b: An open large-scale dataset for training
  next generation image-text models}.
\newblock \bibinfo{journal}{\emph{Advances in neural information processing
  systems}}  \bibinfo{volume}{35} (\bibinfo{year}{2022}),
  \bibinfo{pages}{25278--25294}.
\newblock


\bibitem[Schuhmann et~al\mbox{.}(2021)]%
        {schuhmann2021laion}
\bibfield{author}{\bibinfo{person}{Christoph Schuhmann},
  \bibinfo{person}{Richard Vencu}, \bibinfo{person}{Romain Beaumont},
  \bibinfo{person}{Robert Kaczmarczyk}, \bibinfo{person}{Clayton Mullis},
  \bibinfo{person}{Aarush Katta}, \bibinfo{person}{Theo Coombes},
  \bibinfo{person}{Jenia Jitsev}, {and} \bibinfo{person}{Aran Komatsuzaki}.}
  \bibinfo{year}{2021}\natexlab{}.
\newblock \showarticletitle{Laion-400m: Open dataset of clip-filtered 400
  million image-text pairs}.
\newblock \bibinfo{journal}{\emph{arXiv preprint arXiv:2111.02114}}
  (\bibinfo{year}{2021}).
\newblock


\bibitem[Shi et~al\mbox{.}(2024)]%
        {shi2024instantbooth}
\bibfield{author}{\bibinfo{person}{Jing Shi}, \bibinfo{person}{Wei Xiong},
  \bibinfo{person}{Zhe Lin}, {and} \bibinfo{person}{Hyun~Joon Jung}.}
  \bibinfo{year}{2024}\natexlab{}.
\newblock \showarticletitle{Instantbooth: Personalized text-to-image generation
  without test-time finetuning}. In \bibinfo{booktitle}{\emph{Proceedings of
  the IEEE/CVF conference on computer vision and pattern recognition}}.
  \bibinfo{pages}{8543--8552}.
\newblock


\bibitem[Skalse et~al\mbox{.}(2022)]%
        {skalse2022defining}
\bibfield{author}{\bibinfo{person}{Joar Skalse}, \bibinfo{person}{Nikolaus
  Howe}, \bibinfo{person}{Dmitrii Krasheninnikov}, {and} \bibinfo{person}{David
  Krueger}.} \bibinfo{year}{2022}\natexlab{}.
\newblock \showarticletitle{Defining and characterizing reward gaming}.
\newblock \bibinfo{journal}{\emph{Advances in Neural Information Processing
  Systems}}  \bibinfo{volume}{35} (\bibinfo{year}{2022}),
  \bibinfo{pages}{9460--9471}.
\newblock


\bibitem[Song et~al\mbox{.}(2020)]%
        {song2020denoising}
\bibfield{author}{\bibinfo{person}{Jiaming Song}, \bibinfo{person}{Chenlin
  Meng}, {and} \bibinfo{person}{Stefano Ermon}.}
  \bibinfo{year}{2020}\natexlab{}.
\newblock \showarticletitle{Denoising Diffusion Implicit Models}. In
  \bibinfo{booktitle}{\emph{International Conference on Learning
  Representations}}.
\newblock


\bibitem[Tewel et~al\mbox{.}(2023)]%
        {tewel2023key}
\bibfield{author}{\bibinfo{person}{Yoad Tewel}, \bibinfo{person}{Rinon Gal},
  \bibinfo{person}{Gal Chechik}, {and} \bibinfo{person}{Yuval Atzmon}.}
  \bibinfo{year}{2023}\natexlab{}.
\newblock \showarticletitle{Key-locked rank one editing for text-to-image
  personalization}. In \bibinfo{booktitle}{\emph{ACM SIGGRAPH 2023 conference
  proceedings}}. \bibinfo{pages}{1--11}.
\newblock


\bibitem[Wallace et~al\mbox{.}(2024)]%
        {wallace2024diffusion}
\bibfield{author}{\bibinfo{person}{Bram Wallace}, \bibinfo{person}{Meihua
  Dang}, \bibinfo{person}{Rafael Rafailov}, \bibinfo{person}{Linqi Zhou},
  \bibinfo{person}{Aaron Lou}, \bibinfo{person}{Senthil Purushwalkam},
  \bibinfo{person}{Stefano Ermon}, \bibinfo{person}{Caiming Xiong},
  \bibinfo{person}{Shafiq Joty}, {and} \bibinfo{person}{Nikhil Naik}.}
  \bibinfo{year}{2024}\natexlab{}.
\newblock \showarticletitle{Diffusion model alignment using direct preference
  optimization}. In \bibinfo{booktitle}{\emph{Proceedings of the IEEE/CVF
  Conference on Computer Vision and Pattern Recognition}}.
  \bibinfo{pages}{8228--8238}.
\newblock


\bibitem[Wang et~al\mbox{.}(2024)]%
        {wang2024high}
\bibfield{author}{\bibinfo{person}{Yibin Wang}, \bibinfo{person}{Weizhong
  Zhang}, \bibinfo{person}{Jianwei Zheng}, {and} \bibinfo{person}{Cheng Jin}.}
  \bibinfo{year}{2024}\natexlab{}.
\newblock \showarticletitle{High-fidelity person-centric subject-to-image
  synthesis}. In \bibinfo{booktitle}{\emph{Proceedings of the IEEE/CVF
  Conference on Computer Vision and Pattern Recognition}}.
  \bibinfo{pages}{7675--7684}.
\newblock


\bibitem[Wei et~al\mbox{.}(2023)]%
        {wei2023elite}
\bibfield{author}{\bibinfo{person}{Yuxiang Wei}, \bibinfo{person}{Yabo Zhang},
  \bibinfo{person}{Zhilong Ji}, \bibinfo{person}{Jinfeng Bai},
  \bibinfo{person}{Lei Zhang}, {and} \bibinfo{person}{Wangmeng Zuo}.}
  \bibinfo{year}{2023}\natexlab{}.
\newblock \showarticletitle{Elite: Encoding visual concepts into textual
  embeddings for customized text-to-image generation}. In
  \bibinfo{booktitle}{\emph{Proceedings of the IEEE/CVF International
  Conference on Computer Vision}}. \bibinfo{pages}{15943--15953}.
\newblock


\bibitem[Xiao et~al\mbox{.}(2024)]%
        {xiao2024fastcomposer}
\bibfield{author}{\bibinfo{person}{Guangxuan Xiao}, \bibinfo{person}{Tianwei
  Yin}, \bibinfo{person}{William~T Freeman}, \bibinfo{person}{Fr{\'e}do
  Durand}, {and} \bibinfo{person}{Song Han}.} \bibinfo{year}{2024}\natexlab{}.
\newblock \showarticletitle{Fastcomposer: Tuning-free multi-subject image
  generation with localized attention}.
\newblock \bibinfo{journal}{\emph{International Journal of Computer Vision}}
  (\bibinfo{year}{2024}), \bibinfo{pages}{1--20}.
\newblock


\bibitem[Ye et~al\mbox{.}(2023)]%
        {ye2023ip}
\bibfield{author}{\bibinfo{person}{Hu Ye}, \bibinfo{person}{Jun Zhang},
  \bibinfo{person}{Sibo Liu}, \bibinfo{person}{Xiao Han}, {and}
  \bibinfo{person}{Wei Yang}.} \bibinfo{year}{2023}\natexlab{}.
\newblock \showarticletitle{Ip-adapter: Text compatible image prompt adapter
  for text-to-image diffusion models}.
\newblock \bibinfo{journal}{\emph{arXiv preprint arXiv:2308.06721}}
  (\bibinfo{year}{2023}).
\newblock


\bibitem[Ye et~al\mbox{.}(2024)]%
        {ye2024mplug}
\bibfield{author}{\bibinfo{person}{Qinghao Ye}, \bibinfo{person}{Haiyang Xu},
  \bibinfo{person}{Jiabo Ye}, \bibinfo{person}{Ming Yan},
  \bibinfo{person}{Anwen Hu}, \bibinfo{person}{Haowei Liu}, \bibinfo{person}{Qi
  Qian}, \bibinfo{person}{Ji Zhang}, {and} \bibinfo{person}{Fei Huang}.}
  \bibinfo{year}{2024}\natexlab{}.
\newblock \showarticletitle{mplug-owl2: Revolutionizing multi-modal large
  language model with modality collaboration}. In
  \bibinfo{booktitle}{\emph{Proceedings of the ieee/cvf conference on computer
  vision and pattern recognition}}. \bibinfo{pages}{13040--13051}.
\newblock


\bibitem[Zhang et~al\mbox{.}(2023)]%
        {zhang2023adding}
\bibfield{author}{\bibinfo{person}{Lvmin Zhang}, \bibinfo{person}{Anyi Rao},
  {and} \bibinfo{person}{Maneesh Agrawala}.} \bibinfo{year}{2023}\natexlab{}.
\newblock \showarticletitle{Adding conditional control to text-to-image
  diffusion models}. In \bibinfo{booktitle}{\emph{Proceedings of the IEEE/CVF
  international conference on computer vision}}. \bibinfo{pages}{3836--3847}.
\newblock


\bibitem[Zhang et~al\mbox{.}(2024)]%
        {zhang2024ssr}
\bibfield{author}{\bibinfo{person}{Yuxuan Zhang}, \bibinfo{person}{Yiren Song},
  \bibinfo{person}{Jiaming Liu}, \bibinfo{person}{Rui Wang},
  \bibinfo{person}{Jinpeng Yu}, \bibinfo{person}{Hao Tang},
  \bibinfo{person}{Huaxia Li}, \bibinfo{person}{Xu Tang}, \bibinfo{person}{Yao
  Hu}, \bibinfo{person}{Han Pan}, {et~al\mbox{.}}}
  \bibinfo{year}{2024}\natexlab{}.
\newblock \showarticletitle{Ssr-encoder: Encoding selective subject
  representation for subject-driven generation}. In
  \bibinfo{booktitle}{\emph{Proceedings of the IEEE/CVF Conference on Computer
  Vision and Pattern Recognition}}. \bibinfo{pages}{8069--8078}.
\newblock


\end{thebibliography}


\newpage

\end{document}